\documentclass[sigconf, screen, nonacm]{acmart} %

\setcopyright{none}
\renewcommand\footnotetextcopyrightpermission[1]{} %
\makeatletter
\patchcmd{\@mkauthors@iii}{\lineskip=1pc\relax}{\lineskip=0.65pc\relax}{}{}
\patchcmd{\@mkteasers}{\par\bigskip\bgroup}{\par\smallskip\bgroup}{}{}
\makeatother

\usepackage{CJKutf8}
\usepackage{enumitem} 
\usepackage{subcaption}
\usepackage{graphicx}
\usepackage[table]{xcolor}
\usepackage{tabularx}
\usepackage{pifont}
\usepackage{listings}

\newcommand{\cmark}{\textcolor{green!60!black}{\ding{51}}}
\newcommand{\xmark}{\textcolor{red!70!black}{\ding{55}}}

\lstdefinestyle{promptstyle}{
  basicstyle=\ttfamily\footnotesize,
  breaklines=true,
  breakatwhitespace=true,
  columns=fullflexible,
  keepspaces=true,
  xleftmargin=1em,
  xrightmargin=1em,
  frame=single,
  framesep=4pt,
  rulecolor=\color{black!30},
  backgroundcolor=\color{gray!5},
  showstringspaces=false,
  upquote=true,
}
\AtBeginDocument{%
  }

\begin{document}

\title{TRACE-Bench: Decomposing and Diagnosing Multi-Reference Image Generation}

\author{Haoran Wang}
\authornote{Equal contribution. \textsuperscript{\ensuremath{\dagger}}Project lead. \textsuperscript{\ensuremath{\ddagger}}Corresponding author.}
\email{a.museum@sjtu.edu.cn}
\affiliation[obeypunctuation=true]{%
  \institution{Shanghai Jiao Tong University}
  \city{}
  \country{}}

\author{Chaofan Ma}
\authornotemark[1]
\email{chaofanma@sjtu.edu.cn}
\affiliation[obeypunctuation=true]{%
  \institution{Shanghai Jiao Tong University}
  \city{}
  \country{}}

\author{Ran Yi\texorpdfstring{\textsuperscript{\ensuremath{\dagger}}}{}}
\email{ranyi@sjtu.edu.cn}
\affiliation[obeypunctuation=true]{%
  \institution{Shanghai Jiao Tong University}
  \city{}
  \country{}}

\author{Lizhuang Ma\texorpdfstring{\textsuperscript{\ensuremath{\ddagger}}}{}}
\email{ma-lz@cs.sjtu.edu.cn}
\affiliation[obeypunctuation=true]{%
  \institution{Shanghai Jiao Tong University}
  \city{}
  \country{}}

\renewcommand{\shortauthors}{Wang et al.}

\begin{teaserfigure}
  \centering
  \includegraphics[width=0.93\textwidth,keepaspectratio]{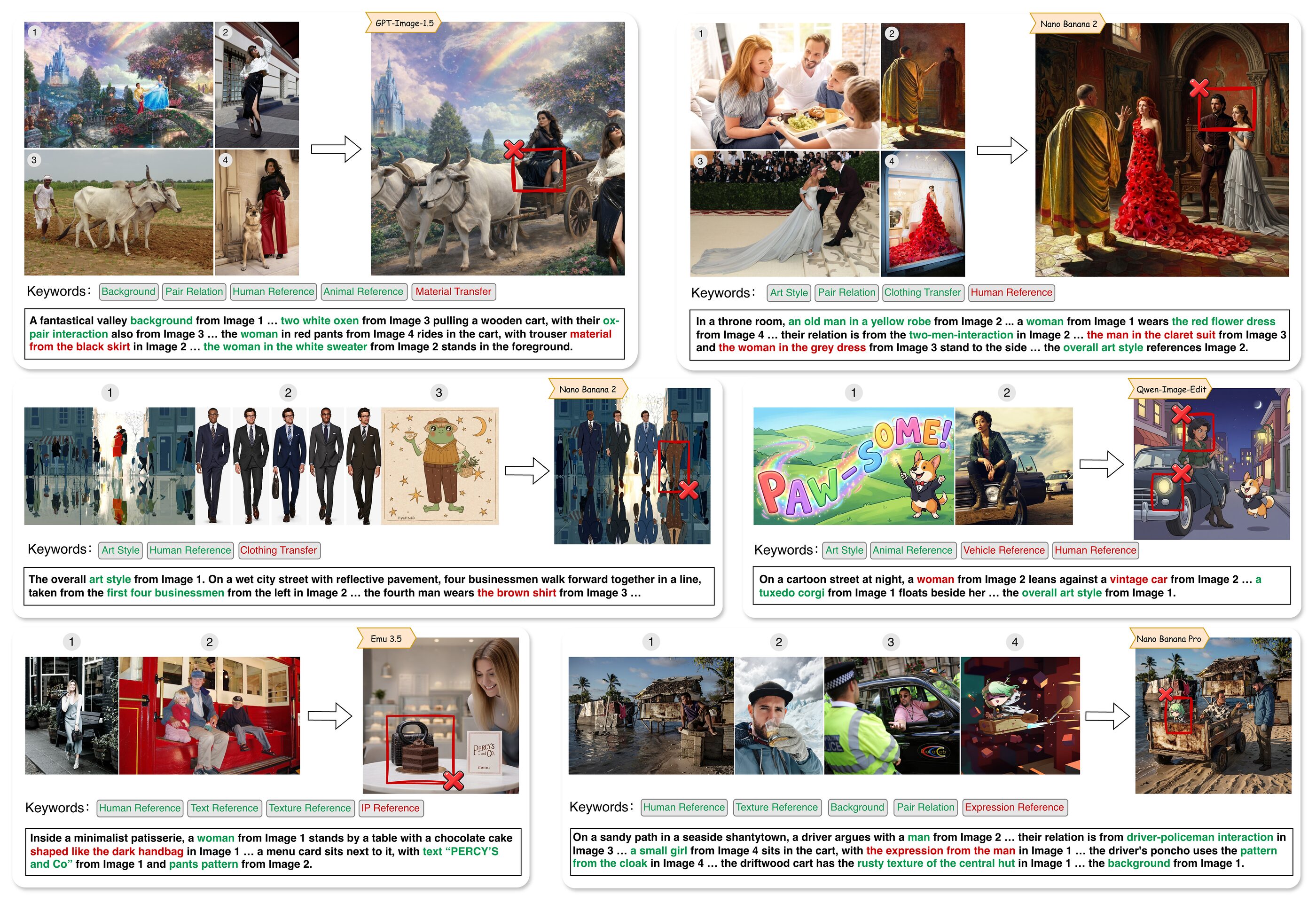}
  \caption{Representative TRACE-Bench cases. Green/red tags indicate satisfied/failed requirements; red boxes localize failures.}
  \Description{Representative multi-reference image-generation cases with reference images, prompts, generated results, and operator-aligned success and failure annotations.}
  \label{fig:teaser}
\end{teaserfigure}

\begin{abstract}
Despite recent advances in unified multimodal models for multi-reference image generation, existing benchmarks remain organized around predefined \textit{task types} (e.g., ``subject composition''), which are ill-suited to this combinatorial setting and lead to fragmented coverage, uncontrolled complexity, and little diagnostic value.
Recognizing that diverse multi-reference tasks share a common set of atomic operations, we adopt a \textit{capability-oriented} perspective and formalize four operators: Anchor~($f$), Disentangle~($g$), Apply~($\oplus$), and Compose~($C$).
Any multi-reference prompt can then be represented as a compositional formula over these operators, whose structural complexity is quantified by the number of operator slots.
Building on this formulation, we construct \textbf{TRACE-Bench}, comprising approximately 1,600 evaluation cases across slot counts 1--8, built from 631 formula templates and around 4,000 reference images spanning diverse artistic styles and real-world subjects.
The formula structure directly drives an \textit{operator-aligned evaluation} protocol for per-capability scoring and a \textit{diagnostic tree} analysis for recursive failure localization.
Evaluating 9 leading models reveals insights invisible to holistic scoring: the primary bottleneck lies in disentanglement~($g$) and attribute binding~($\oplus$) rather than scene-level composition~($C$), with even the best model scoring only 0.74 on attribute fidelity.
Project page: \href{https://amuseum-whr.github.io/TraceBench}{https://amuseum-whr.github.io/TraceBench}
\end{abstract}

\maketitle
\pagestyle{plain}
\thispagestyle{plain}

\begin{figure*}[!t]
    \centering
    \includegraphics[width=0.94\linewidth]{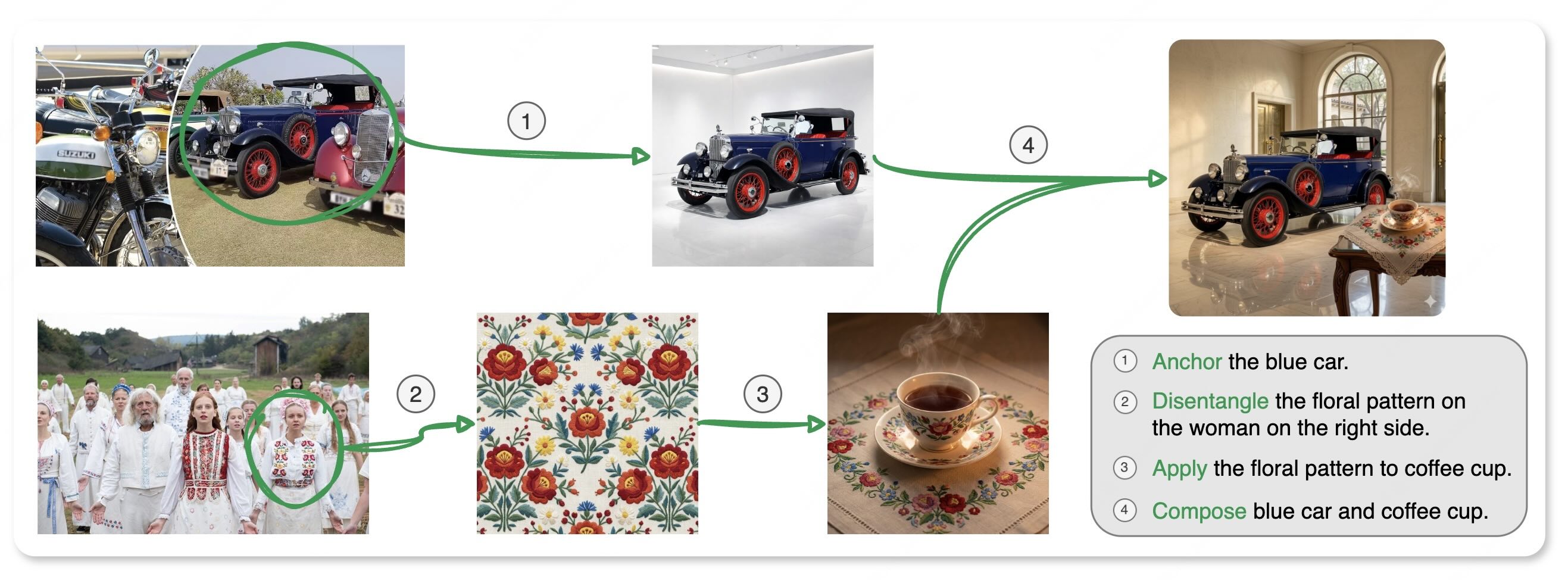}
    \caption{A multi-reference request in TRACE-Bench is progressively resolved through \textit{Anchor}, \textit{Disentangle}, \textit{Apply}, and \textit{Compose}.}
    \Description{A four-stage illustration showing how Anchor, Disentangle, Apply, and Compose progressively construct a multi-reference image-generation result.}
    \label{fig:core_capabilities}
\end{figure*}

\section{Introduction}

Text-to-image generation~\cite{rombach2022high,esser2024scaling} has achieved remarkable success, yet text alone is often insufficient to convey precise visual details.
This limitation has motivated reference-based image generation, which allows users to ground outputs in user-provided visual content.
A particularly challenging extension is multi-reference generation, where models must jointly condition on multiple visual elements—a capability essential to real-world workflows such as virtual try-on, group photo composition, and multi-source creative design.
Recent models, both proprietary (GPT-Image-1.5~\cite{openai2025gptimage15}, Nano Banana 2~\cite{google2026nanobanana2}) and open-source (OmniGen2~\cite{wu2025omnigen2}, Emu3.5~\cite{Cui2025Emu35NM}, Qwen-Image-Edit~\cite{Wu2025QwenImageTR}), have demonstrated strong capabilities in following instructions that combine entities, attributes, and styles from different sources. 

On the evaluation side, while existing benchmarks have extensively addressed text-to-image alignment~\cite{huang2023t2icompbench,ghosh2023geneval} and single-image editing~\cite{wu2025kris,zhao2025risebench}, evaluation of multi-reference generation remains in its early stages.
Recent efforts like MultiBanana~\cite{oshima2025multibanana}, MICON-Bench~\cite{wu2026miconbench}, and MacroBench~\cite{chen2026macro} have pioneered this direction.
However, while these works cover more complex reference settings, they do not fundamentally rethink the underlying evaluation structure.
Following the design of earlier generation and editing benchmarks, they still organize test cases around predefined \textit{task types} (\textit{e.g.}, ``object composition'').
In the combinatorial setting of multi-reference generation, this task-oriented organization exposes three critical limitations.
\textbf{(1) Incomplete coverage}: predefined task categories cannot scale to the full combinatorial space of practical multi-reference usage.
\textbf{(2) No failure diagnosis}: holistic task-level scoring cannot pinpoint which specific capability is responsible for a failure; for example, if a model fails to generate a person wearing a referenced outfit, a single score cannot reveal whether the failure stems from misidentifying the person, incorrectly extracting the outfit, or wrongly binding the outfit to the target.
\textbf{(3) Uncontrolled complexity}: without a unifying structure that formally characterizes each case, it is difficult to systematically control or compare structural complexity across scenarios.

These limitations motivate us to rethink the evaluation of multi-reference generation from a \textbf{capability-oriented perspective}. 
Our key observation is that seemingly diverse multi-reference generation tasks share a common set of \textbf{atomic operations}. 
Consider a prompt such as ``\emph{Generate a scene containing the blue car from [Image~1] and a coffee cup decorated with the floral pattern worn by the woman on the right in [Image~2], with the cup placed on a table to the lower right of the car.}'' 
As illustrated in Fig.~\ref{fig:core_capabilities}, fulfilling this prompt requires the model to first \textbf{\textit{anchor}} the intended blue car from a visually cluttered reference image while preserving its identity-defining characteristics; 
\textbf{\textit{disentangle}} the floral pattern worn by the woman on the right, separating this transferable attribute from its original carrier; 
\textbf{\textit{apply}} the disentangled pattern to a new carrier (the coffee cup), binding the extracted attribute to a target instance; 
and finally \textbf{\textit{compose}} the anchored car together with the modified cup into a coherent scene, with the cup placed on a table to the lower right of the car. 
We formalize these as four capability operators: Anchor~($f$), Disentangle~($g$), Apply~($\oplus$), and Compose~($C$). 
This capability-oriented formulation naturally overcomes the three limitations of task-oriented evaluation identified above.
\textbf{(1)} Any complex multi-reference prompt can be expressed as a compositional formula over these operators, elegantly covering the infinite combinatorial space of real-world usage without needing ad-hoc task labels.
\textbf{(2)} This formulation enables operator-aligned evaluation: rather than assigning a holistic score, we can precisely diagnose which capability (\textit{e.g.}, identity preservation in $f$ or attribute exclusivity in $\oplus$) caused a failure.
\textbf{(3)} The reference-conditioned structural complexity of any test case can be rigorously quantified and systematically controlled by the number of operator slots in its underlying formula.
Moreover, common applications such as virtual try-on and group photo layout emerge naturally as specific instantiations of this compositional framework, demonstrating its expressiveness and practical coverage.

Building on this formulation, we construct \textbf{TRACE-Bench}, a capability-oriented benchmark for multi-reference image generation.
\textbf{TRACE-Bench} comprises approximately 1,600 cases built from 631 formula templates involving around 4,000 reference images, with slot counts ranging from 1 to 8 for systematic control over structural complexity.
We collect reference images from multiple complementary sources spanning diverse artistic styles and real-world subjects, and apply structured tagging to extract entities and attributes in a fine-grained hierarchy for formula-driven prompt construction.
Each sampled formula template is realized as a natural-language prompt by a vision-language model (VLM), and paired with an operator-aligned evaluation checklist scored by a VLM judge.
As summarized in Table~\ref{tab:eval_dims}, this checklist associates each capability with a corresponding evaluation dimension and a set of fine-grained criteria.
Fig.~\ref{fig:teaser} shows representative benchmark cases and their operator-aligned evaluation results.
Beyond case-level scoring, \textbf{TRACE-Bench} further supports diagnostic tree analysis, which recursively decomposes complex failure cases into simpler sub-cases to identify the responsible source of failure.

\begin{table}[t]
\centering
\caption{Operator-aligned evaluation dimensions.}
\label{tab:eval_dims}
\small
\setlength{\tabcolsep}{3pt}
\begin{tabular}{@{}p{1.8cm}p{1.7cm}p{3.9cm}@{}}
\toprule
\textbf{Operator} & \textbf{Dimension} & \textbf{Criteria} \\
\midrule
$f:$ Anchor & Identity & Existence; appearance consistency with the reference \\
$g:$ Disentangle & Attribute Fidelity & Presence; consistency with the reference source \\
$\oplus:$ Apply & Binding & Carrier integrity; attribute exclusivity; natural integration \\
$C:$ Compose & Composition & Coexistence; relation satisfaction; spatial coherence; no duplication or leakage \\
\bottomrule
\end{tabular}
\end{table}

We evaluate 9 leading proprietary and open-source models on \textbf{TRACE-Bench}.
Our operator-level analysis yields two key insights not captured by conventional holistic scoring.
First, the primary bottleneck in current models lies in attribute disentanglement ($g$) and attribute binding ($\oplus$) rather than scene-level composition ($C$), indicating that precise reference transfer remains substantially harder than plausible scene arrangement. 
Second, anchor difficulty is driven more by the number of entities in the reference image than by formula slot count, suggesting that reference-image clutter, not task structure, is the dominant source of error.

Our contributions are summarized as follows:
\begin{itemize}[leftmargin=*, itemsep=2pt, topsep=3pt, parsep=0pt, partopsep=0pt]
    \item We propose a capability-oriented formulation that decomposes reference-based image generation into four atomic operators, with a compositional formula for systematically characterizing diverse multi-reference settings.
    \item We construct \textbf{TRACE-Bench}, a benchmark of approximately 1,600 cases with slot-based complexity control (slot 1--8), built from 631 formula templates and around 4,000 reference images.
    \item We design an operator-aligned evaluation protocol and a diagnostic tree analysis method that enable fine-grained, per-capability assessment and failure localization.
    \item We benchmark 9 leading models and reveal that the primary bottleneck lies in attribute disentanglement and attribute binding rather than scene-level composition, among other insights invisible to holistic scoring.
\end{itemize}

\section{Related Work}

\subsection{Reference-Based Image Generation}

Early text-to-image models~\cite{rombach2022high, ramesh2022hierarchical} generate images purely from text prompts, offering limited control over fine-grained visual details. To address this, reference-based methods enable users to condition generation on visual examples. DreamBooth~\cite{ruiz2023dreambooth} and Textual Inversion~\cite{gal2023textualinversion} fine-tune or learn embeddings from a small set of reference images to capture subject identity. ControlNet~\cite{zhang2023controlnet} and IP-Adapter~\cite{ye2023ipadapter} introduce auxiliary conditioning branches that accept spatial or semantic signals from a single reference image without fine-tuning.

While these methods achieve strong results in single-reference settings, they are not inherently designed for multi-reference generation, where inputs from several images must be jointly processed.
Early multi-subject methods such as Custom Diffusion~\cite{kumari2023customdiffusion} and FastComposer~\cite{xiao2023fastcomposer} extend personalization to multiple concepts but still require per-subject optimization or rely on localized attention mechanisms tied to a fixed set of subjects.
More recently, unified multimodal models have emerged that natively support interleaved image-text inputs, enabling flexible multi-reference generation within a single forward pass.
Proprietary systems such as GPT-Image-1.5~\cite{openai2025gptimage15} and the Nano Banana series~\cite{google2025nanobanana,google2025nanobananapro,google2026nanobanana2} demonstrate strong multi-reference capabilities.
On the open-source side, OmniGen2~\cite{wu2025omnigen2}, BAGEL~\cite{deng2025bagel}, and Emu3.5~\cite{Cui2025Emu35NM} adopt unified architectures that jointly handle understanding and generation;
FLUX.1 Kontext~\cite{bfl2025kontext}, Qwen-Image-Edit~\cite{Wu2025QwenImageTR}, and FireRed Image Edit~\cite{firered2026rededit} support instruction-based generation and editing conditioned on reference images;
DeepGen~\cite{wang2026deepgen} and UniReason~\cite{wang2026unireason} emphasize lightweight and reasoning-centered unification of generation and editing.
As these models grow increasingly capable, how to rigorously evaluate their multi-reference abilities at a fine-grained capability level remains an open challenge.

\subsection{Multi-Reference Generation Benchmarks}

Extensive benchmarks already exist for text-to-image alignment~\cite{huang2023t2icompbench,ghosh2023geneval,hu2024dpgbench,li2025t2icorebench} and single-image editing~\cite{geditbench2026v2,wu2025kris,zhao2025risebench}, yet evaluation of multi-reference generation remains nascent.
MultiBanana~\cite{oshima2025multibanana} scales evaluation to 8 reference images and introduces difficulty factors such as domain mismatch and rare concepts.
MICON-Bench~\cite{wu2026miconbench} defines six compositional tasks for multi-image context generation.
OmniContext~\cite{wu2025omnigen2} introduces 8 task categories for in-context generation, though its scope is limited to relatively simple subject-centric compositions.
MacroBench~\cite{chen2026macro} provides 4,000 samples across four task dimensions with up to 10 references.

Despite expanding the scope and scale of multi-reference evaluation, these benchmarks share a common design philosophy: organizing test cases around predefined task types or surface-level difficulty factors, and assessing results with holistic scores or coarse metrics such as FID~\cite{heusel2017fid} and CLIP similarity~\cite{radford2021learning}.
Although recent evaluation practices have advanced towards VLM-as-a-judge protocols~\cite{ku2024visionlanguagemodels} and structured checklist questions~\cite{li2025t2icorebench,wei2025tiifbench}, the underlying benchmark design still lacks two critical properties: it cannot systematically control evaluation complexity across structurally different cases, and it cannot localize failures to specific capability dimensions or distinguish standalone weaknesses from cross-reference interference.
\textbf{TRACE-Bench} addresses both gaps by decomposing multi-reference generation into four atomic capability operators and using their compositional structure as the unified basis for benchmark construction, operator-aligned evaluation, and diagnostic failure analysis.

\section{TRACE-Bench}

\label{sec:method}

\begin{figure*}[!t]
    \centering
    \includegraphics[width=0.93\linewidth]{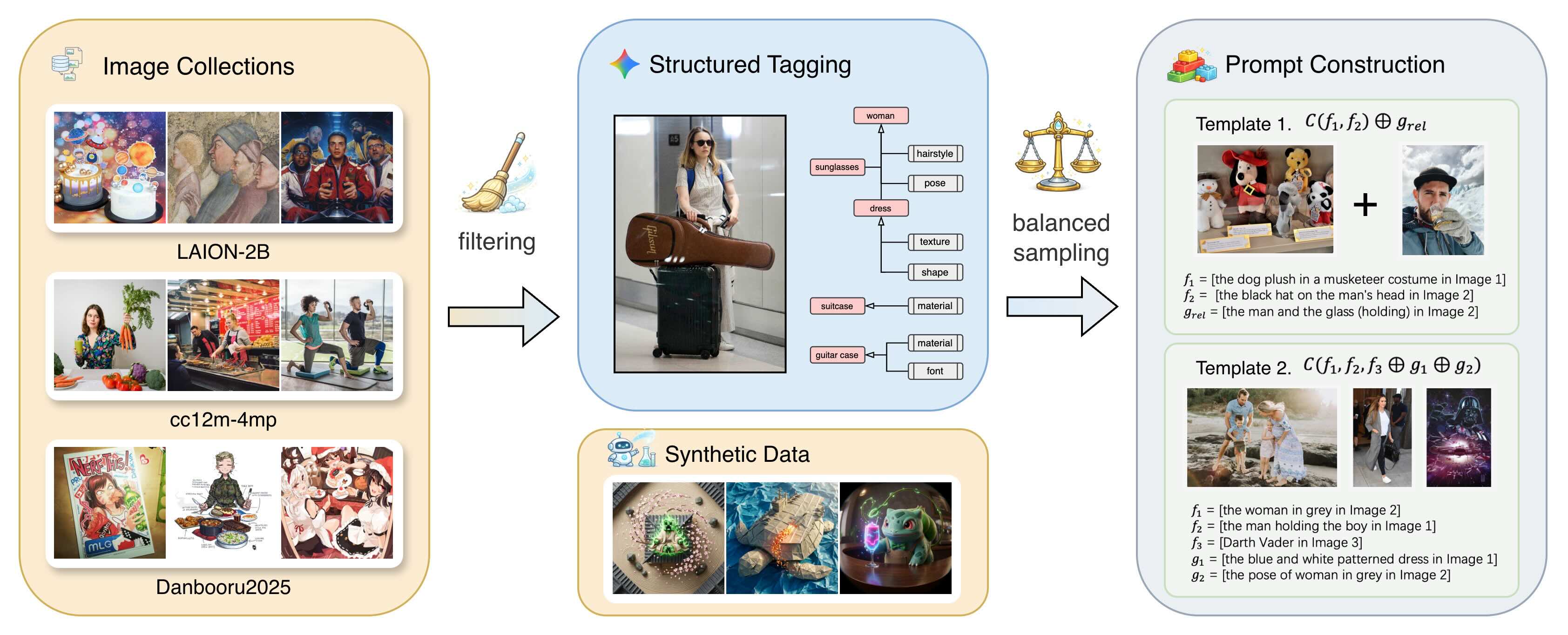}
    \caption{Overview of the benchmark construction pipeline. Candidate images are first collected and filtered from multiple sources, then annotated through structured tagging. The tagged pool is then balanced through source-wise sampling and augmented with synthetic data. It is subsequently used for formula-template sampling and prompt construction.}
    \Description{A pipeline diagram showing image collection and filtering, structured tagging, balanced sampling and synthetic augmentation, formula-template sampling, and prompt construction.}
    \label{fig:data_pipeline}
\end{figure*}

\subsection{Overview}
\label{sec:overview}

Evaluating multi-reference image generation requires both diverse test cases and a structured representation of what each case demands.
TRACE-Bench therefore couples capability-oriented benchmark construction with operator-aligned evaluation.
Sec.~\ref{sec:capability} introduces the four core operators, and Sec.~\ref{sec:symbolic} defines the compositional formula that structures each prompt.
Secs.~\ref{sec:construction} and~\ref{sec:evaluation} then detail benchmark construction and operator-aligned evaluation.

\subsection{Capability Decomposition}
\label{sec:capability}

A key difficulty in benchmarking multi-reference image generation is that holistic scores can obscure fine-grained task failures~\cite{ghosh2023geneval}, while general perceptual quality can diverge from source-conditioned validity~\cite{rong2026h2rbench}, making it hard to identify whether a failure comes from poor image generation or incorrect use of the references. To trace a model's reference ability more explicitly, we decompose it into four core capabilities: \emph{Anchor}, \emph{Disentangle}, \emph{Apply}, and \emph{Compose}. Let $I$ denote a reference image, $e$ an entity in a reference image, $\mathcal{E}$ a referenced entity set, and $a$ a referenced attribute. We use $T_e$ to denote an entity specified in the text prompt.

\begin{itemize}[leftmargin=*, itemsep=3pt]
    \item \textbf{Anchor} $f(I, e)$: locating a specific entity $e$ in reference image $I$ and preserving its identity-defining visual information in the generated image. For example, $f(I_1, \texttt{person})$ denotes the person in Image~1 as the target entity to preserve.

    \item \textbf{Disentangle} $g(I, \mathcal{E}, a)$: extracting a referenced attribute $a$ from entity set $\mathcal{E}$ in image $I$, while decoupling it from irrelevant properties. Here, $\mathcal{E}$ may contain a single entity or multiple entities, depending on the attribute type. For example, in Case 1 of Fig.~\ref{fig:case_study}, $g(I_1, \{\texttt{robe}\}, \texttt{pattern})$ denotes extracting \emph{only} the decorative pattern on the robe in Image~1, while discarding the robe's shape and the identity of the camel wizard wearing it.

    \item \textbf{Apply} $\oplus$: binding a disentangled attribute to a designated entity. The designated entity may be either an anchored entity $f(\cdot)$ from reference images or an entity specified in the text prompt. For example, $T_e \oplus g_1$ denotes applying the extracted attribute $g_1$ to the text-described entity $T_e$ (e.g., applying the running pose of the man from Image 1 to a robot).

    \item \textbf{Compose} $C(\cdot)$: arranging multiple referenced or text-specified contents into a coherent scene, optionally under additional relational constraints. For example, $C(f_1,\; T_e \oplus g_1) \oplus g_{\text{rel}}$ denotes composing the anchored entity $f_1$ with a text-described entity $T_e$ modified by $g_1$, while further enforcing a referenced relation $g_{\text{rel}} = g(I_3, \{e_i, e_j\}, a_{\text{rel}})$ between them. If the desired relation is specified in the text prompt rather than referenced from an image, we denote it by $T_{\text{rel}}$.
\end{itemize}

These four operators form the atomic capability space of multi-reference generation, but a real prompt typically nests several of them at once. We next organize such nested structure into a compositional formula.

\subsection{Formula Composition}
\label{sec:symbolic}

The four operators of Sec.~\ref{sec:capability} give us the atomic vocabulary, but describing how a real case combines them purely in natural language leaves the underlying structure implicit: which references are involved, how they interact, and how difficult the overall case is are all buried inside free-form text. We therefore represent each prompt's reference-conditioned part as a compositional formula over these operators. Making this structure symbolic turns each case into a shared backbone that the rest of TRACE-Bench builds on: its operator terms can be systematically enumerated to form a template space of diverse cases, the number of reference-dependent terms provides a controllable measure of structural complexity, and each operator instance seeds an evaluation question aligned with the corresponding capability. The formula only captures reference-conditioned content, namely which entities are anchored, which attributes are disentangled, where they are applied, and how the resulting contents are composed. Text-only descriptions that do not depend on any reference stay in natural language.

We organize the formula from local to global with three levels: entity expressions (single target objects) $\rightarrow$ scene expressions (compositions of entities) $\rightarrow$ the complete prompt formula (entire reference-conditioned structure).

\noindent\textbf{Entity Expressions.}
An entity expression $E$ describes a target subject in the generated image. It may be an anchored reference entity $f$, a text-specified carrier modified by a disentangled attribute $T_e \oplus g$, or an existing entity expression further augmented with additional attributes, written as $E \oplus g$. Thus, the entity level answers what each generated subject is and which reference-derived attributes are bound to it.

\noindent\textbf{Scene Expressions.}
A scene expression $S$ composes multiple entity expressions through $C(E_1, E_2, \ldots, E_n)$, optionally together with a relation term. The relation may be specified by text, $T_{\text{rel}}$, or extracted from a reference image, $g_{\text{rel}}$. When a referenced relation applies only to a subset of entities, we represent that subset as a nested sub-scene such as $C(E_i, E_j)\oplus g_{\text{rel}}$, and then compose it into the larger scene. The scene level therefore answers how the target objects coexist and interact.

\begin{figure*}[t]
    \centering
    \includegraphics[width=0.91\linewidth]{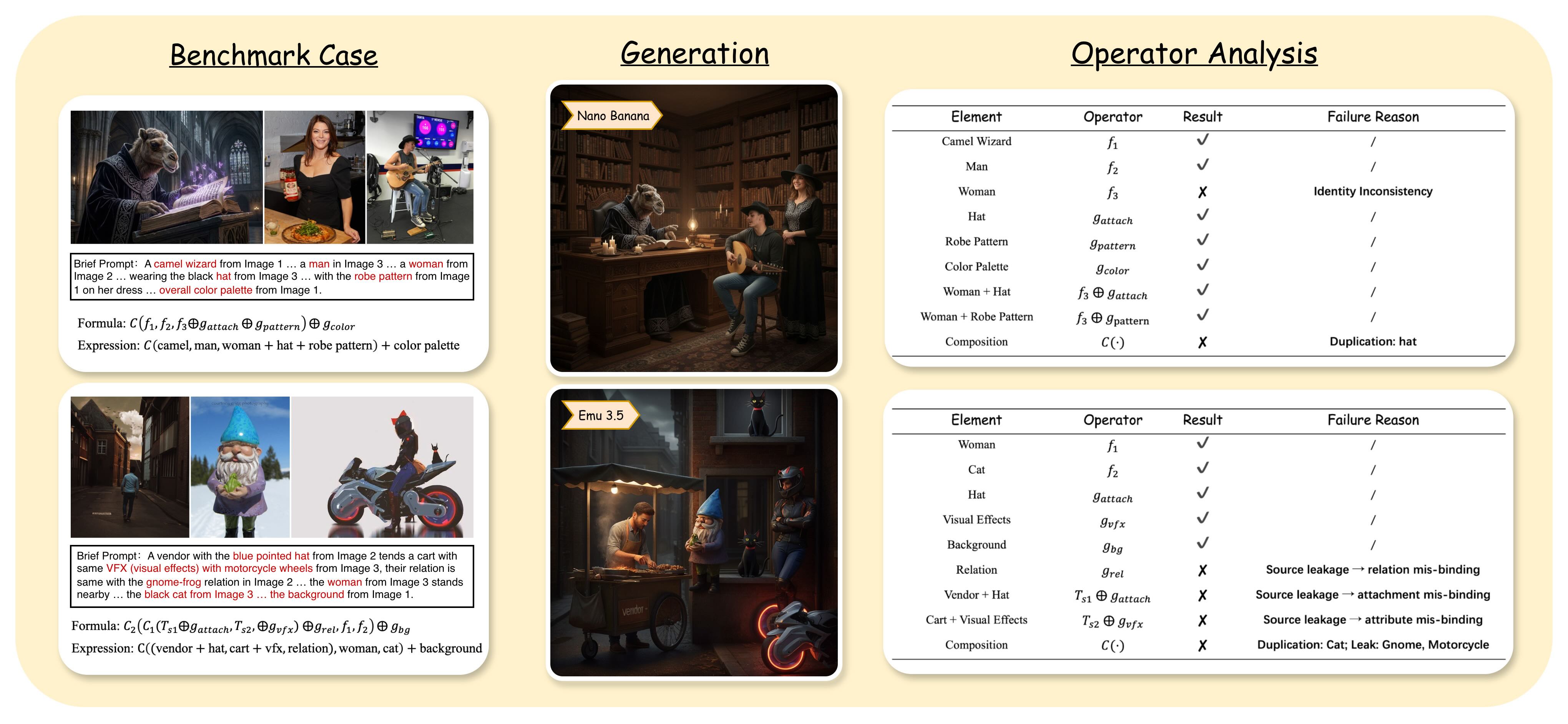}
    \caption{Representative operator-aligned evaluation examples from TRACE-Bench. Each row shows one benchmark case with its references, brief prompt, compositional formula, readable expression, generated result, and operator-level analysis. The examples illustrate how formula terms are mapped to capability-specific checks and how failures such as identity mismatch, duplication, and leakage-induced mis-binding can be localized.}
    \Description{Two benchmark examples showing reference images, prompts, compositional formulas, generated results, and operator-level evaluations that identify identity, duplication, and binding failures.}
    \label{fig:case_study}
\end{figure*}

\noindent\textbf{Complete Prompt Formulas.}
A complete prompt formula $F$ further augments the scene expression $S$ with optional global reference conditions, such as style, lighting, layout, or color tone, denoted by $g_{\text{global}}$. For example,
$F = C\bigl(C(f_1,\, T_e \oplus g_1) \oplus g_{\mathrm{rel}},\, f_2 \oplus g_2 \oplus g_3\bigr) \oplus g_{\mathrm{global}}$.
Here, the inner composition groups two entities ($f_1$ and $T_e \oplus g_1$) under $g_{\mathrm{rel}}$, the outer composition combines this sub-scene with another attribute-modified entity ($f_2 \oplus g_2 \oplus g_3$), and the final global term applies a scene-level reference condition. We use this formula as the canonical structure for benchmark construction and evaluation. For readability, Fig.~\ref{fig:case_study} also presents the same formula as a noun-based expression while preserving the underlying structure.

\subsection{Benchmark Construction}
\label{sec:construction}

The remaining question is how to instantiate many diverse formulas into concrete benchmark prompts. To this end, we design a structured construction pipeline that combines multi-source image curation, structured tagging, balanced sampling, formula-template sampling, and prompt generation. Fig.~\ref{fig:data_pipeline} illustrates the overall pipeline, and we describe its stages below.

\noindent\textbf{Image Collection and Filtering.}
We collect candidate reference images from three complementary sources: Danbooru2025~\cite{trojblue_danbooru2025_metadata}, a large-scale anime and illustration dataset with rich stylistic diversity and well-defined character designs; LAION-2B-en-Aesthetics~\cite{laion2b_en_aesthetic}, a subset of LAION-5B covering diverse internet image-text data; and cc12m-4mp-realistic~\cite{opendiffusionai_cc12m_4mp_realistic}, a human-focused subset of Conceptual Captions that strengthens coverage of real human subjects. We then apply source-specific filtering: for Danbooru, we retain only images with $\texttt{score} > 26.15$ that satisfy the safety filter; for LAION, we retain only samples with $\texttt{aesthetic} > 6.5$ and $\texttt{similarity} > 30$. After filtering, we sample approximately 50,000 candidate images spanning diverse artistic styles and visual themes.

\noindent\textbf{Structured Tagging.}
Each image is annotated by a Gemini-2.5-pro-based tagging pipeline that summarizes foreground entities and their associated attributes in a structured form. Entities are assigned category labels from a predefined ontology, including \emph{Human}, \emph{Animal}, \emph{Object}, \emph{Food}, \emph{Clothing}, \emph{Transportation}, \emph{Structure}, and \emph{Text}. Attributes are organized into four layers---\emph{Appearance}, \emph{Form}, \emph{Dynamics}, and \emph{Global}---each further divided into finer-grained subcategories. In addition, each entity is associated with a grounding phrase for localizing it in the image, and each attribute is accompanied by a short textual description to facilitate downstream prompt construction.

\noindent\textbf{Balanced Sampling.}
Since category and attribute distributions differ substantially across sources, we perform source-wise balanced sampling to improve long-tail coverage. For each source, each candidate image is represented by a feature vector $\mathbf{x}_i$ derived from the tagging results. We then greedily select samples according to
\begin{equation}
\label{eq:balanced_sampling}
i^\star=\arg\max_i \; \mathbf{x}_i^\top \mathbf{w}^{(t)}, \qquad
w_j^{(t)}=\frac{1}{1+c_j^{(t)}},
\end{equation}
where $\mathbf{c}^{(t)}$ denotes the accumulated feature counts of the selected subset at iteration $t$. This helps to yield a more balanced reference pool while preserving visual diversity. Balanced sampling selects approximately 4,000 images (about 8\% of the candidate pool), after which full manual image-quality inspection retains 3,839 images. We further augment the pool with about 200 synthetic samples generated by Nano Banana Pro to supplement rare cases.

\noindent\textbf{Structural Complexity Control.}
Each benchmark prompt is associated with a compositional formula, and we use its \emph{slot count} as a controllable measure of structural complexity:
\begin{equation}
\label{eq:slot_count}
\text{slot}(F) := |f| + |g|,
\end{equation}
where $|f|$ and $|g|$ count anchored and disentangled terms in $F$, respectively; slot count measures formula structure rather than fully determining case difficulty. \textbf{TRACE-Bench} covers slots 1--8, from simple single-reference to highly compositional cases. For example, formula
$C(f_1,\; T_e \oplus g_1 \oplus g_2)\oplus g_{\mathrm{global}}$
is a slot-4 case.

\noindent\textbf{Template Sampling.}
We first construct each benchmark instance as a formula template composed of the operators defined above. In the standard case, both $f$ and $g$ are sampled from the structured tagging results. To better cover practical applications, we further introduce two special designs in which an anchored instance $f$ is used as an attribute-like reference term $g$: \emph{attachment reference} ($g_{\text{attach}}$), where a referenced instance serves as an attachable component of another entity, and \emph{IP-style reference} ($g_{\text{ip}}$), where the holistic design identity is transferred to another entity. Across all slot levels, we sample from the template space under controlled distributions.

\noindent\textbf{Prompt Generation.}
Each sampled template is paired with tagged reference images and fed into a VLM (Gemini-2.5-Pro) through a customized prompting interface, which realizes it as a natural-language prompt. We require that the resulting prompt describe a coherent scene, clearly bind each reference to a specific target, and use every referenced image at least once. For quality control, we first use GPT-5.4 to filter out 4.3\% of the constructed prompts. We then manually inspect the remaining cases and remove another 9\%. For each benchmark prompt, we additionally construct a text-only counterpart in which all image-referenced descriptions are replaced by textual ones, providing a no-reference baseline.

\noindent\textbf{Benchmark Statistics.}
For the general benchmark, we construct \textbf{180} cases for each slot level, and further include several application-specific cases. In total, the benchmark contains approximately \textbf{1,600} cases, built from \textbf{631} distinct formula templates and involving around \textbf{4,000} reference images.

\begin{table*}[t]
    \centering
    \caption{Overall results on \textbf{TRACE-Bench} averaged over slots 1--8. Avg. denotes the mean of the four operator-aligned metrics. Best results are in \textbf{bold} and second-best results are highlighted with a gray background.}
    \label{tab:main_results}
    \small
    \begin{tabular}{lcccccc}
    \toprule
    \textbf{Model} & \textbf{Anchor ($\mathbf{f}$)} & \textbf{Disentangle ($\mathbf{g}$)} & \textbf{Apply ($\boldsymbol{\oplus}$)} & \textbf{Compose ($\mathbf{C}$)} & \textbf{Avg.} & \textbf{CLIP Sim} \\
    \midrule
    GPT-Image-1.5          & \cellcolor{gray!15}0.7649 & 0.6890 & 0.7541 & \textbf{0.9259} & 0.8118 & \textbf{0.2969} \\
    Nano Banana            & 0.7650 & 0.6786 & 0.7631 & 0.8975 & 0.7981 & 0.2867 \\
    Nano Banana 2          & \textbf{0.7724} & \textbf{0.7384} & \textbf{0.7989} & 0.9100 & \textbf{0.8205} & 0.2944 \\
    Nano Banana Pro        & 0.7488 & \cellcolor{gray!15}0.7148 & \cellcolor{gray!15}0.7869 & \cellcolor{gray!15}0.9214 & \cellcolor{gray!15}0.8172 & \cellcolor{gray!15}0.2962 \\
    \midrule
    Emu3.5                 & 0.6587 & 0.4982 & 0.5434 & 0.7871 & 0.6561 & 0.2917 \\
    FireRed Image Edit 1.1~\cite{firered2026rededit,firered2026github}
                            & 0.6348 & 0.4703 & 0.4258 & 0.7218 & 0.5889 & 0.2603 \\
    Qwen-Image-Edit-2509   & 0.5210 & 0.3755 & 0.3282 & 0.7627 & 0.5483 & 0.2742 \\
    Qwen-Image-Edit-2511   & 0.6009 & 0.4097 & 0.3742 & 0.7776 & 0.5858 & 0.2758 \\
    OmniGen2~\cite{wu2025omnigen2}
                            & 0.5635 & 0.3719 & 0.3195 & 0.7070 & 0.5348 & 0.2693 \\
    \bottomrule
    \end{tabular}
    \end{table*}

\subsection{Evaluation Protocol}
\label{sec:evaluation}

\noindent\textbf{Operator-Aligned Question Generation.}
Our formula-based construction explicitly grounds each referenced term to its source image, enabling evaluation questions to be derived automatically from the formula structure. Rather than assigning a single holistic score to the generated result, we decompose evaluation into operator-aligned question sets following Table~\ref{tab:eval_dims}.

Specifically, anchor ($\mathbf{f}$) evaluates entity existence and consistency; disentangle ($\mathbf{g}$) evaluates attribute existence and consistency; apply ($\boldsymbol{\oplus}$) evaluates binding correctness and integration quality; and compose ($\mathbf{C}$) evaluates compositional coherence and the absence of anomalies such as duplication or leakage.

\noindent\textbf{VLM-Based Judging.}
A representative evaluation example is shown in Fig.~\ref{fig:case_study}. For each benchmark sample, we assess referenced entities with $f$, referenced attributes with $g$, attribute application with $\oplus$, and scene composition with $C$. All questions are scored in a binary manner by a VLM judge (Gemini-2.5-Pro), which receives the reference images, the generated image, and the operator-aligned question set, and outputs a pass/fail decision for each question.

\noindent\textbf{Evaluation Metrics.}
For each operator instance, we instantiate a set of fine-grained evaluation questions that assess complementary aspects of the same capability. Their scores are normalized such that the aggregate contribution of each operator instance equals 1. Case-level scores are then obtained by aggregating the normalized scores across all operator instances in the case.

\noindent\textbf{Diagnostic Tree Analysis.}
Diagnostic-tree decomposition reverses the local-to-global formula construction introduced in Sec.~\ref{sec:symbolic}. Starting from the complete formula at the root, we progressively remove full-prompt-level terms to recover the underlying scene expression, separate the scene into entity expressions, and simplify multi-attribute bindings until each leaf contains a single anchored entity $f$ or an atomic attribute transfer $T_e \oplus g$. Whenever reference-conditioned content is removed, it is replaced with a corresponding text-only description to preserve the original prompt context. The resulting sub-cases form a tree. Evaluating and comparing its nodes allows us to localize the source of a failure observed at the root, as illustrated by the decomposition of a slot-4 formula in Fig.~\ref{fig:diagnostic_tree}. The complete construction rules are detailed in Appendix~\ref{sec:diagnostic_tree_rules}.

\begin{figure}[h]
    \centering
    \includegraphics[width=0.99\linewidth]{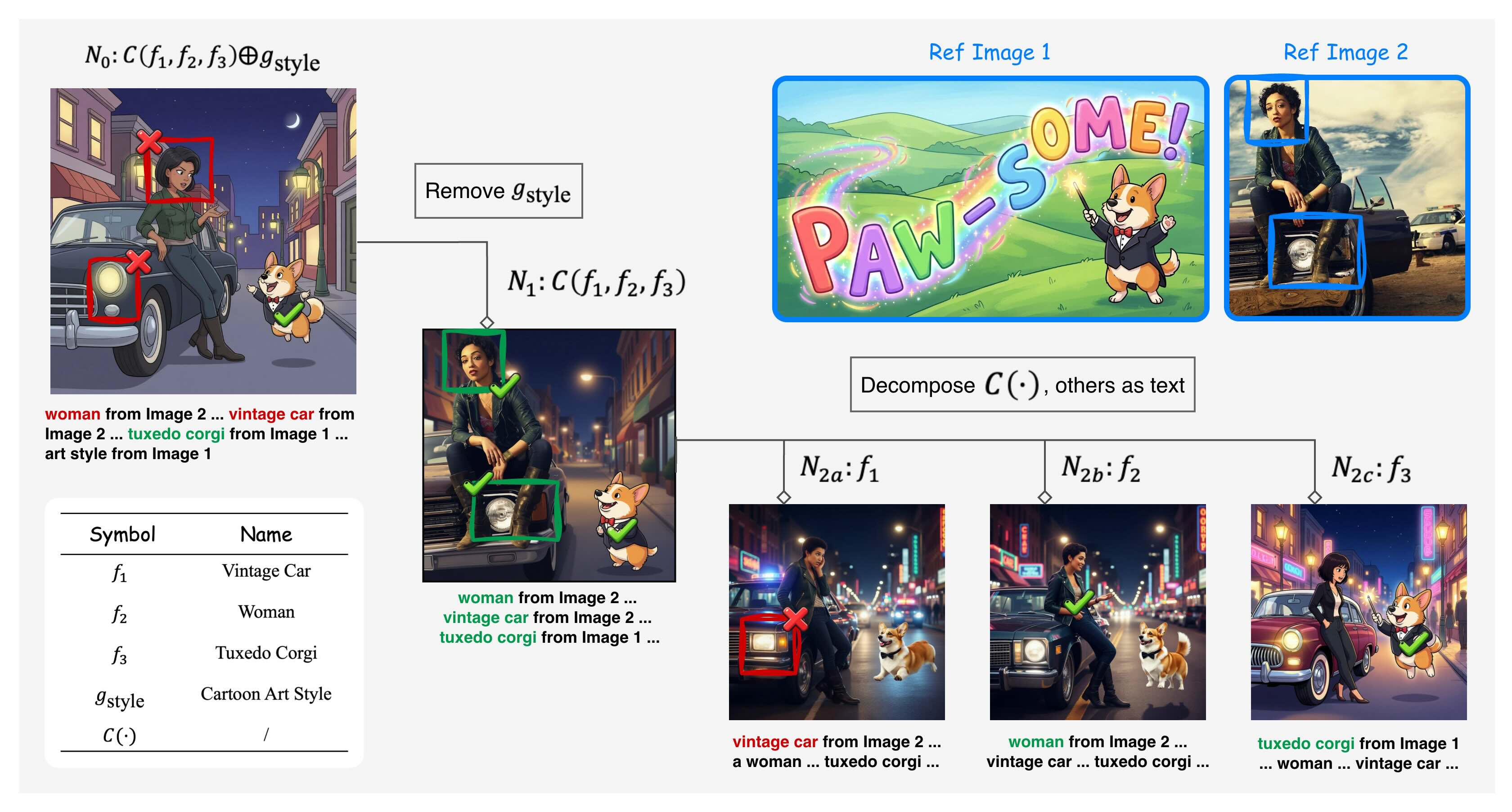}
    \caption{Representative diagnostic tree analysis case.}
    \Description{A diagnostic tree that recursively decomposes a multi-reference generation case into simpler sub-cases to identify where reference-conditioned failures arise.}
    \label{fig:diagnostic_tree}
\end{figure}

\section{Experiments}

\subsection{Experimental Setup}

\noindent\textbf{Baselines.}
We compare against 9 representative baselines, including 4 proprietary models and 5 open-source models. The proprietary models are GPT-Image-1.5~\cite{openai2025gptimage15}, Nano Banana~\cite{google2025nanobanana}, Nano Banana Pro~\cite{google2025nanobananapro}, and Nano Banana 2~\cite{google2026nanobanana2}. The open-source models are Emu3.5~\cite{Cui2025Emu35NM}, FireRed Image Edit 1.1~\cite{firered2026rededit,firered2026github}, Qwen-Image-Edit~\cite{Wu2025QwenImageTR,qwenimagegithub} with two released versions (2509 and 2511), and OmniGen2~\cite{wu2025omnigen2}.

\noindent\textbf{Evaluation Setup.}
Unless otherwise specified, all reported scores are computed on the full benchmark. Operator-aligned evaluation uses Gemini-2.5-Pro as the VLM judge. We retain a text-only prompt for each case and additionally report text-image similarity computed by CLIP ViT-L/14~\cite{radford2021learning} as a supplementary metric.

\subsection{Overall Evaluation}

Using the operator-aligned checklist in Sec.~\ref{sec:evaluation}, we evaluate all baselines on \textbf{TRACE-Bench}. Table~\ref{tab:main_results} reports the overall results. Proprietary models consistently outperform open-source baselines, with Nano Banana 2 achieving the best average score. However, the task is still far from solved. Since each operator score is normalized to an ideal value of 1, even the best model remains well below saturation: 0.7724 on anchor, 0.7384 on disentangle, 0.7989 on apply, and 0.9100 on compose.

The largest gaps appear on disentangle and apply, namely $g$ and $\oplus$. Even the strongest models remain far below 1 on these dimensions, showing that correct attribute extraction and target assignment are still the main bottlenecks. Composition is relatively stronger: GPT-Image-1.5 reaches 0.9259 on $C$. Anchor is also more stable, but the best score is still only 0.7724.

Among open-source models, Emu3.5 performs best overall. Qwen-Image-Edit-2511 improves over Qwen-Image-Edit-2509 on all four metrics. Still, all open-source baselines remain substantially behind the leading proprietary systems, especially on $g$ and $\oplus$. CLIP similarity follows a similar ranking trend, but it is less sensitive to whether the referenced content is transferred to the correct target.

\noindent\textbf{Qualitative Case Analysis.}
Fig.~\ref{fig:case_study} shows two representative cases from \textbf{TRACE-Bench}: one generated by Nano Banana and the other by Emu3.5. These examples illustrate the value of such fine-grained evaluation. In the Nano Banana case, the generated image appears plausible overall, but the analysis reveals an identity mismatch for the referenced woman and a composition error caused by hat duplication. In the Emu3.5 case, the referenced woman, cat, hat, visual effects, and background are all present in the scene, but the referenced relation and attribute application both fail because of source leakage. These examples show that \textbf{TRACE-Bench} can localize specific failure modes rather than collapsing them into a single overall judgment.

\begin{figure}[h]
    \centering
    \includegraphics[width=0.95\linewidth]{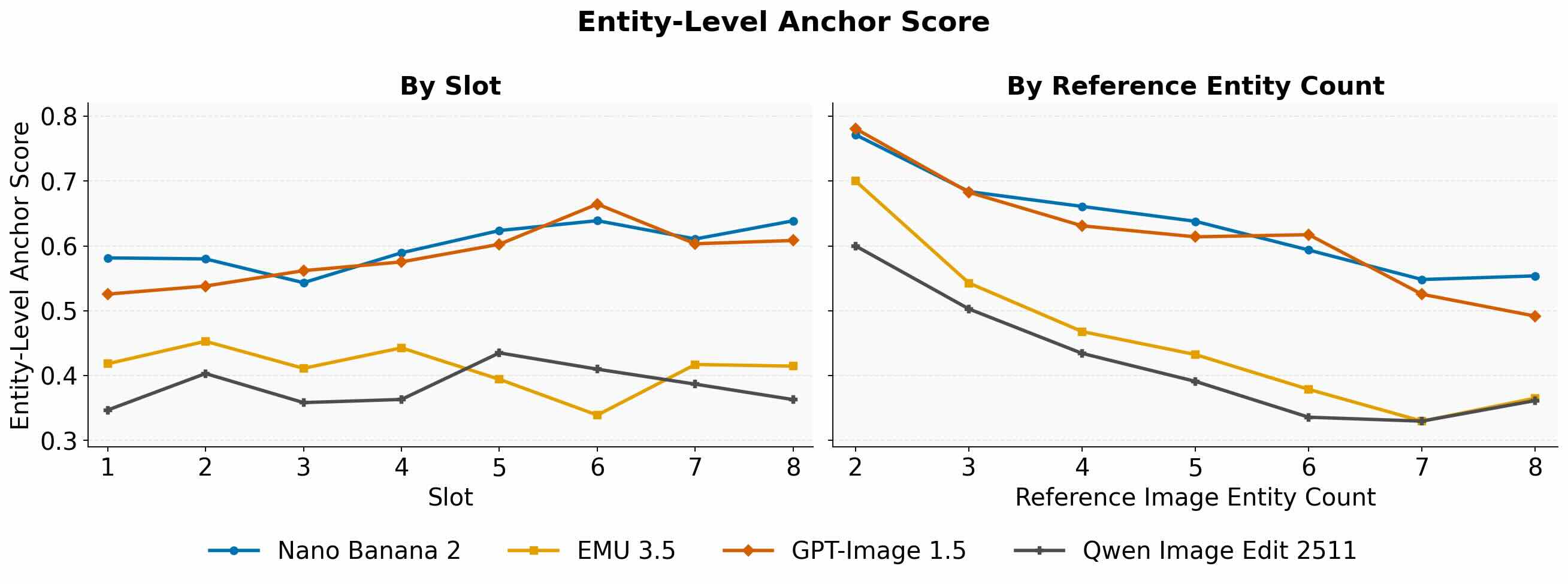}
    \caption{Anchor performance versus template slot count (left) and reference-image entity count (right).}
    \Description{Two line charts comparing anchor scores across template slot counts and across numbers of entities in the reference image for several evaluated models.}
    \label{fig:anchor_analysis}
\end{figure}

\begin{figure*}[!t]
    \centering
    \includegraphics[width=0.999\linewidth]{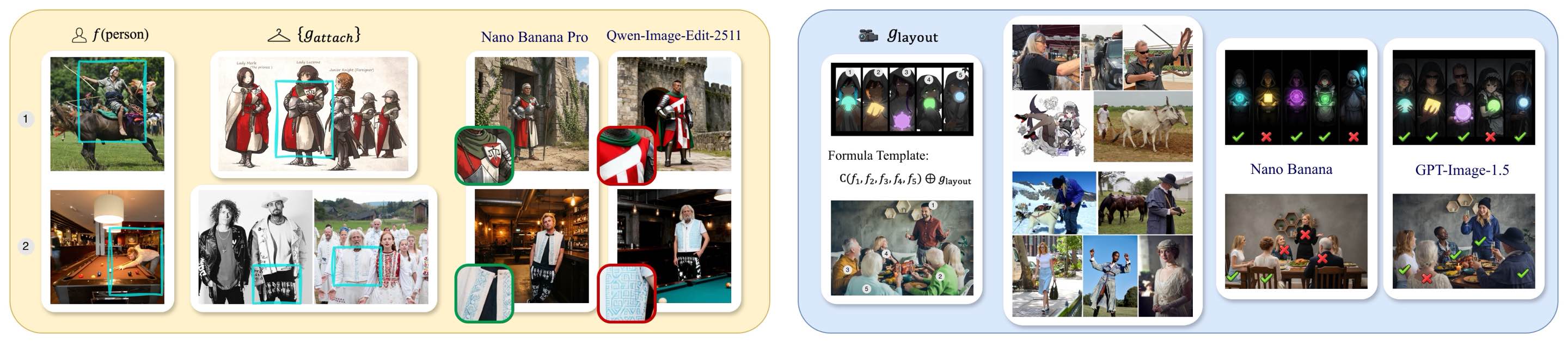}
    \caption{Formula abstractions and representative results for virtual try-on (left) and group-photo layout (right).}
    \Description{Examples of virtual try-on and group-photo layout showing their formula abstractions, reference images, generated results, and success or failure markers.}
    \label{fig:app_cases}
\end{figure*}

\noindent\textbf{Anchor Under Different Difficulty Factors.}
To illustrate how different sources of difficulty can affect a specific capability, we take \emph{anchor} as an example and compare its performance by slot level and by the number of entities in the reference image. As shown in Fig.~\ref{fig:anchor_analysis}, anchor performance varies only weakly across slot levels, but declines more clearly as the reference image contains more entities. This suggests that, for anchor, reference-image complexity is a more direct source of difficulty than slot count alone. More broadly, it highlights the value of operator-level analysis in \textbf{TRACE-Bench}: even when two cases have similar overall structural complexity, they may differ substantially in the difficulty of a specific capability.

\subsection{Application-Oriented Analysis}

During formula-template sampling and prompt construction, we observe that many common applications can be naturally expressed within our compositional framework. Rather than defining them as separate task types, we treat them as particular instantiations of the same core operators. Figure~\ref{fig:app_cases} illustrates two representative examples: virtual try-on and group-photo layout.

\noindent\textbf{Virtual Try-On.}
Virtual try-on binds one or more clothing-related references to a target person:
\begin{equation}
\label{eq:virtual_tryon}
f(\texttt{person}) \oplus g_{\text{attach},1} \oplus g_{\text{attach},2} \oplus \cdots.
\end{equation}
This pattern covers different numbers and types of garments. In the left example of Fig.~\ref{fig:app_cases}, the successful result preserves both the target person and the referenced clothing, whereas the failed result transfers incorrect garment attributes.

\noindent\textbf{Group Photo Layout.}
Group-photo layout composes multiple anchored subjects under a shared layout reference:
\begin{equation}
\label{eq:group_layout}
C(f_1, f_2, \ldots, f_n) \oplus g_{\text{layout}}.
\end{equation}
The same pattern extends to different group sizes and spatial arrangements. As shown on the right, satisfying the shared layout may cause identity loss or subject duplication.

\begin{table}[h]
\centering
\caption{Distribution of diagnostic outcomes across 200 cases.}
\label{tab:diagnostic_sources}
\small
\setlength{\tabcolsep}{3pt}
\begin{tabular}{@{}lccccc@{}}
\toprule
\textbf{Outcome / Source} & $f$ & $g$ & $\oplus$ & $C$ & \textbf{Overall} \\
\midrule
Stable Success       & 27.9 & 27.6 & 38.9 & 45.2 & 33.7 \\
Persistent Failure   & 4.7  & 13.2 & 7.3  & 12.9 & 9.8  \\
Global Reference     & 14.0 & 10.5 & 9.1  & \textbf{29.0} & 13.7 \\
Joint Composition    & \textbf{48.8} & \textbf{43.4} & \textbf{41.1} & 9.7 & \textbf{38.5} \\
Relation / Attribute & 4.7  & 5.3  & 3.6  & 3.2 & 4.4 \\
\bottomrule
\end{tabular}
\end{table}

\subsection{Diagnostic Tree Analysis}
\label{sec:diagnostic}

In Fig.~\ref{fig:teaser}, the Qwen-Image-Edit-2511 example fails on the woman and the vintage car. To better understand these errors, we further analyze this case with the diagnostic tree in Fig.~\ref{fig:diagnostic_tree}, where we decompose the original formula into simpler sub-cases and evaluate the model on each node.

Two distinct patterns emerge. For \(f_2\), the woman is correct in \(N_1\) and \(N_{2b}\), but becomes inconsistent in \(N_0\). This indicates that $f_2$ itself is not the problem; rather, identity information is lost when the global style constraint is introduced. For \(f_1\), the vintage car already changes in \(N_{2a}\), showing that this anchor is intrinsically harder. Yet it is preserved in \(N_1\), suggesting that jointly referencing the interacting subject \(f_2\) can reinforce its identity.

This example shows that the diagnostic tree can distinguish two failure sources within the same case: style-induced interference for \(f_2\), and intrinsic anchor difficulty for \(f_1\), which is partially alleviated under composition.

\noindent\textbf{Aggregate Diagnostic Patterns.}
We construct diagnostic trees for 200 Emu3.5 cases, generate an image at every node, and score each node using the operator-aligned evaluation. For each operator instance that fails at the root, we identify the first decomposition step at which it passes and attribute the failure to the reference-conditioned component removed at that step. Instances whose outcomes remain unchanged throughout the tree are categorized as Stable Success or Persistent Failure.

As shown in Table~\ref{tab:diagnostic_sources}, joint-composition interference is the dominant localized source for $f$, $g$, and $\oplus$. This indicates that Emu3.5 often preserves isolated reference content but loses it when multiple reference-conditioned entities are composed. In contrast, failures in $C$ are most frequently localized to global-reference interference (29.0\%), suggesting that global style or scene constraints are a major source of compositional disruption. Overall, most localized failures arise from interactions introduced at higher compositional levels rather than from persistent failure on isolated reference units.

\section{Conclusion}

We presented \textbf{TRACE-Bench}, a capability-oriented benchmark for multi-reference image generation.
Rather than organizing evaluation around predefined task types, we decompose multi-reference generation into four atomic operators (Anchor, Disentangle, Apply, Compose).
Their compositional structure serves as the unified basis for benchmark construction, operator-aligned evaluation, and diagnostic tree analysis.
Evaluation of 9 leading models reveals that the primary bottleneck lies in attribute disentanglement ($g$) and binding ($\oplus$) rather than scene-level composition ($C$), and that diagnostic tree analysis can effectively separate cross-reference interference from standalone capability deficits.
In the future, we hope the capability-oriented formulation can serve not only as an evaluation tool but also as a guide for targeted model improvement.

\begin{acks}
This work was supported by the National Natural Science Foundation of China (No. 62302297, 72192821, 62472282, 62272447, 62472285), the Fundamental Research Funds for the Central Universities (project number: YG2023QNA35), YuCaiKe [2023] Project Number: 231111310300.
\end{acks}

\bibliographystyle{ACM-Reference-Format}
\bibliography{sample-base}

\appendix

\begin{figure*}[t]
    \centering
    \includegraphics[width=0.96\linewidth]{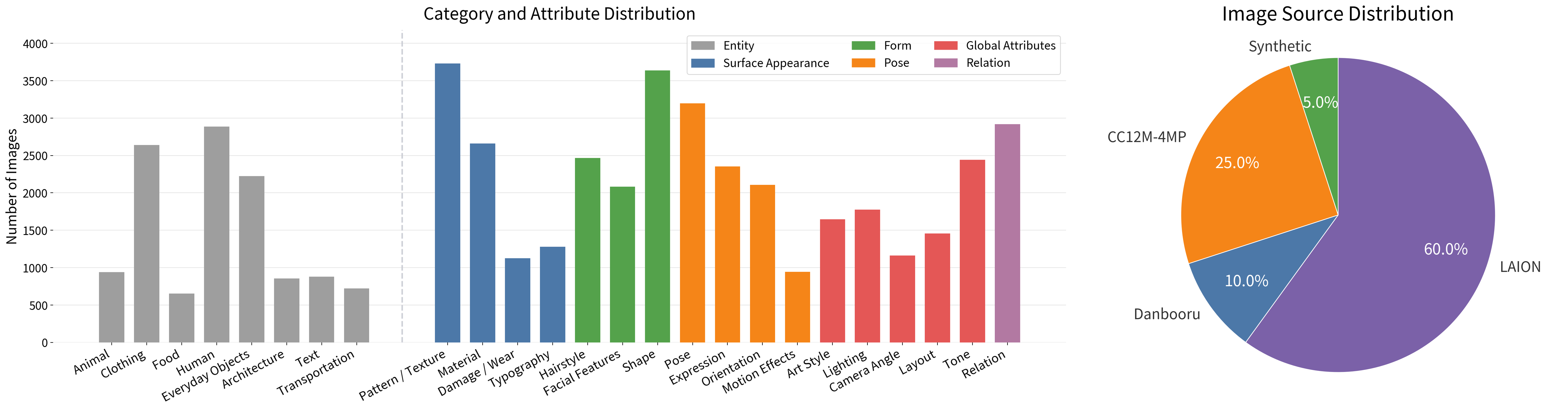}
    \caption{Benchmark statistics after construction. The left panel shows the distributions of semantic categories and controllable attributes, and the right panel shows the composition of image sources.}
    \label{fig:distribution}
\end{figure*}

\section{Task Formulation and Symbolic Representation Details}

\subsection{Comparison with Existing Benchmarks}
\label{sec:comparison_existing_benchmarks}

Compared with existing benchmarks, TRACE-Bench differs in three main aspects, as summarized in Table~\ref{tab:benchmark_comparison}.

\textbf{Capability-oriented decomposition.}
A key difference lies in the principle used to organize the benchmark. Existing benchmarks are mostly decomposed by \emph{task categories}, such as reference-count settings, predefined subtasks, or long-context task types. In contrast, TRACE-Bench is decomposed by \emph{atomic capabilities}, namely Anchor ($f$), Disentangle ($g$), Apply ($\oplus$), and Compose ($C$). This capability-oriented design allows different generation scenarios to be analyzed within a shared capability space, rather than being treated as isolated task categories. In this sense, TRACE-Bench differs not only in benchmark split, but also in the basis on which benchmark cases are constructed.

\textbf{Compositional case construction with aligned evaluation.}
A second key difference is the explicit connection between case construction and evaluation. Existing benchmarks typically define cases through task templates, subtasks, or task-specific settings, while the evaluation protocol is designed as a separate layer. In contrast, TRACE-Bench constructs cases from compositional formulas and uses the same underlying structure to define evaluation targets. During construction, the formula specifies how reference-conditioned contents are combined into a benchmark instance. During evaluation, the same structure determines the operator targets and the corresponding checklist items. As a result, benchmark construction and evaluation are explicitly aligned, which makes the overall pipeline more coherent and also makes the meaning of each evaluation item easier to interpret.

\textbf{Richer reference content and harder grounding.}
A third difference lies in the richness of the reference content and the difficulty of grounding it correctly. Our candidate image pool is designed to contain more diverse and information-rich images, and the tagging system extracts more comprehensive referenceable content from each image. As a result, a single image may provide multiple usable reference contents, rather than serving only as a source for one object or one simple global attribute. This makes both Anchor and Disentangle more challenging: the intended reference may need to be distinguished through language-aligned, attribute-based, or prototype representations~\cite{li2022lseg,xu2022groupvit,ma2022fusioner,ma2023attrseg,yang2024multimodal,zhang2023prototypical} and then grounded from referring descriptions in multi-object scenes~\cite{wang2022cris,yang2024remamber,mao2025safire}, with TRACE-Bench further including same-class distractors. The examples in Fig.~\ref{fig:core_capabilities} illustrate this point: in Steps~1 and~2, the correct reference cannot be identified by a simple noun phrase alone, but instead requires more specific grounding. For this reason, Appendix~\ref{sec:prompt_realization} further provides detailed prompt realization rules for such cases.

\begin{table*}[t]
\centering
\small
\setlength{\tabcolsep}{6pt}
\renewcommand{\arraystretch}{1.12}
\begin{tabular}{lcccc}
\toprule
\textbf{Benchmark} & \textbf{Split by} & \textbf{Case basis} & \textbf{Eval. alignment} & \textbf{Hard grounding} \\
\midrule
OmniContext        & Tasks        & Subtasks               & \xmark & \xmark \\
MultiBanana        & Tasks        & Reference-count tasks  & \xmark & \xmark \\
MICON-Bench        & Tasks        & Task templates         & \xmark & \cmark \\
MacroBench         & Tasks        & Long-context tasks     & \cmark & \xmark \\
TRACE-Bench (ours) & Capabilities & Compositional formulas & \cmark & \cmark \\
\bottomrule
\end{tabular}
\caption{Comparison with existing benchmarks. ``Split by'' indicates the primary principle used to decompose the benchmark. ``Case basis'' summarizes the main basis used to construct individual benchmark cases. ``Eval. alignment'' indicates whether the evaluation protocol is explicitly aligned with the benchmark construction logic. ``Hard grounding'' indicates whether identifying the intended reference content often requires detailed localization descriptions rather than simple noun phrases.}
\label{tab:benchmark_comparison}
\end{table*}

\section{Benchmark Construction Details}

\subsection{Attribute Taxonomy and Tagging Criteria}

Our attribute taxonomy is organized into four levels: \textbf{Appearance}, \textbf{Form}, \textbf{Dynamics}, and \textbf{Global}. A key design consideration is that human and humanoid entities require substantially finer-grained annotation than ordinary objects. Both in practical applications and in perceptual evaluation, users are typically more sensitive to identity and appearance errors on humans than on other categories. As a result, our taxonomy includes several tags that are especially important for human-centered references, such as hairstyle, facial features, and expression.

Appearance attributes describe visible local appearance details of an instance. Specifically, \textbf{Pattern/Texture} refers to repeated or local surface appearance, such as floral prints, stripes, embroidery, or decorative motifs. \textbf{Material} describes what the surface appears to be made of, such as metal, glass, wood, fur, or knitted fabric. \textbf{Damage/Wear} captures visible aging or usage traces, such as scratches, rust, cracks, folds, or worn edges. \textbf{Font/Text Style} is used when the visual identity of text itself is important, including letterform style, stroke shape, and decorative typography.

Form attributes describe relatively stable structural or morphology-related properties of an instance. \textbf{Hairstyle} describes hair-related appearance of human or humanoid entities, including length, curliness, bangs, braids, and overall styling. \textbf{Facial Features} refers to visually recognizable facial characteristics such as beard, makeup, eye shape, nose shape, or other salient facial details. \textbf{Shape} captures the overall geometric or morphological form of an entity, especially for objects, creatures, or clothing silhouettes.

Dynamics attributes capture transient states or motion-related properties. \textbf{Action/Pose} describes the body configuration or ongoing motion of an instance, such as running, sitting, raising one hand, or leaning forward. \textbf{Expression} is mainly used for humans or humanoid characters, and captures facial states such as smiling, frowning, surprise, or anger. \textbf{Orientation/Position} records how the instance is oriented or spatially placed, such as facing left, side view, front-facing, or lying on a surface. \textbf{Motion Effect} captures visible motion-related effects or dynamic cues, such as splashing water, flying sparks, motion trails, or magical glow produced during an action.

Finally, global attributes describe image-level properties that are not naturally attached to a single foreground instance. \textbf{Style} refers to the overall rendering style of the image, such as oil painting, anime, watercolor, or realistic photography. \textbf{Lighting} describes global illumination conditions, such as backlighting, warm indoor light, or strong contrast. \textbf{Camera/Viewpoint} captures the overall photographic perspective, such as close-up, top-down view, side shot, or wide-angle composition. \textbf{Color Tone} describes the overall palette or grading, such as warm-toned, low-saturation, or blue-dominant. \textbf{Layout/Composition} refers to the global arrangement of major scene elements. \textbf{Inter-instance Relation} describes explicit relations among multiple instances, such as holding, standing beside, hugging, or facing each other.

Overall, this taxonomy is designed to balance transferability, perceptual salience, and annotation stability, while remaining aligned with the operator-based formulation used throughout the benchmark.

\subsection{Structured Tagging Format}
\label{sec:structured_tagging_format}

Based on the attribute taxonomy above, we organize the tagging result of each reference image into a structured representation with three top-level fields: \texttt{ent\_list}, \texttt{background}, and \texttt{global\_tag}. Listing~\ref{lst:tagging_schema} shows the overall schema.

The field \texttt{ent\_list} contains the foreground instances selected from the image. Each instance is represented by a grounding-oriented description \texttt{ent\_desc}, a coarse category label \texttt{category}, and a nested attribute dictionary \texttt{tag\_dict}. In \texttt{tag\_dict}, attributes are grouped into three levels: \textbf{appearance attributes}, \textbf{form attributes}, and \textbf{dynamics attributes}. This design keeps each transferable attribute explicitly attached to the instance it belongs to.

\begin{lstlisting}[float=t, caption={Schema of the structured tagging format.}, label={lst:tagging_schema}]
{
  "ent_list": [
    {
      "category": "...",
      "ent_desc": "...",
      "tag_dict": {
        "1_Appearance": {
          "1.1_pattern_texture": ["..."],
          "1.2_material": ["..."],
          "1.3_damage_wear": ["..."],
          "1.4_font": ["..."]
        },
        "2_Form": {
          "2.1_hairstyle": ["..."],
          "2.2_facial_features": ["..."],
          "2.3_shape": ["..."]
        },
        "3_Dynamics": {
          "3.1_action_pose": ["..."],
          "3.2_expression": ["..."],
          "3.3_orientation": ["..."],
          "3.4_motion_effect": ["..."]
        }
      }
    }
  ],
  "background": "...",
  "global_tag": {
    "4.1_style": ["..."],
    "4.2_lighting": ["..."],
    "4.3_camera_viewpoint": ["..."],
    "4.4_layout_sequence": ["..."],
    "4.5_color_tone": ["..."],
    "4.6_inter_instance_relation": ["..."]
  }
}
\end{lstlisting}

Although our formulation separates \textbf{Anchor} and \textbf{Disentangle}, attribute extraction in real images is still naturally instance-based. We therefore represent each image using an ``instance + attached attributes'' format. The field \texttt{background} records visually salient background content that is useful for later prompt construction but is not treated as a foreground instance. The field \texttt{global\_tag} stores image-level attributes, including salient relations among foreground instances when these relations are useful for later case construction.

This structured format follows a simple principle: we first identify meaningful foreground instances, and then attach transferable attributes to them. In this way, the grounding of each attribute remains explicit, and the resulting representation is easier to use in later formula construction.

In addition to ordinary fine-grained attributes, the structured format also supports two special transferable types, namely $g_{\text{attach}}$ and $g_{\text{ip}}$. We place them in this section because they are represented in a more holistic way than standard local tags.

$g_{\text{attach}}$ is used for attachment-like transferable content, such as clothing, accessories, or other attached components whose identity should be preserved as a whole during transfer. For example, when transferring the T-shirt worn by a man in Image~1 to a woman in the target image, the clothing item should remain intact, rather than being decomposed into a few isolated local attributes.

$g_{\text{ip}}$ is used for holistic IP-style transfer. In such cases, the transferable content is not a single local attribute, but the overall design language of an instance. For example, when generating ``a hat in the style of the rabbit police officer in Image~1,'' the reference signal includes the characteristic silhouette, color scheme, and iconic motifs of the original design.

Overall, this structured tagging format serves as an intermediate representation between raw reference images and later formula instantiation. It preserves explicit grounding at the instance level, while also retaining background information and global attributes when needed.

\subsection{Template Construction and Sampling Strategy}
\label{sec:template_sampling}

\begin{table*}[t]
\centering
\footnotesize
\setlength{\tabcolsep}{5pt}
\begin{tabular}{c c}
\toprule
\textbf{Slot} & \textbf{Example formula templates} \\
\midrule
1 &
$T_e \oplus g$; \quad
$f$ \\
\midrule
2 &
$C(f,\, T_e \oplus g)$; \quad
$f \oplus g$; \quad
$T_e \oplus g \oplus g$; \quad
$C(f,\, f)$ \\
\midrule
3 &
$f \oplus g \oplus g$; \quad
$C(f,\, T_e \oplus g) \oplus g_{\mathrm{global}}$; \quad
$C(f,\, f) \oplus g_{\mathrm{global}}$; \quad
$C(f,\, T_e \oplus g,\, T_e \oplus g)$ \\
\midrule
4 &
$C(T_e \oplus g,\, f,\, f) \oplus g_{\mathrm{global}}$; \quad
$C(T_e \oplus g,\, T_e \oplus g,\, f) \oplus g_{\mathrm{global}}$; \quad
$C(f,\, f \oplus g) \oplus g_{\mathrm{global}}$ \\
\midrule
5 &
$C(T_e \oplus g,\, f,\, f,\, f) \oplus g_{\mathrm{global}}$; \quad
$C(C(T_e \oplus g,\, f) \oplus g_{\mathrm{rel}},\, f) \oplus g_{\mathrm{global}}$; \quad
$C(C(T_e \oplus g,\, T_e \oplus g) \oplus g_{\mathrm{rel}},\, f) \oplus g_{\mathrm{global}}$ \\
\midrule
6 &
$C(C(f,\, f) \oplus g_{\mathrm{rel}},\, T_e \oplus g \oplus g) \oplus g_{\mathrm{global}}$; \quad
$C(T_e \oplus g,\, f,\, f \oplus g \oplus g) \oplus g_{\mathrm{global}}$ \\
\midrule
7 &
$C(C(f,\, f,\, f) \oplus g_{\mathrm{rel}},\, f \oplus g) \oplus g_{\mathrm{global}}$; \quad
$C(C(T_e \oplus g,\, f,\, f) \oplus g_{\mathrm{rel}},\, f \oplus g) \oplus g_{\mathrm{global}}$ \\
\midrule
8 &
$C(C(T_e \oplus g,\, f \oplus g) \oplus g_{\mathrm{rel}},\, T_e \oplus g,\, f \oplus g) \oplus g_{\mathrm{global}}$; \quad
$C(C(f,\, f) \oplus g_{\mathrm{rel}},\, T_e \oplus g,\, T_e \oplus g,\, f) \oplus g_{\mathrm{global}} \oplus g_{\mathrm{global}}$ \\
\bottomrule
\end{tabular}
\caption{Example formula templates across different slot levels.}
\label{tab:example_formula_by_slot}
\end{table*}

This section provides additional details on how template construction is built upon the local-to-global formula levels introduced in the main paper.

\textbf{Local-to-global formula levels.}
At the entity level, the formula describes individual reference-conditioned targets. At the scene level, multiple targets are composed through $C$, optionally together with relational terms. At the full-prompt level, the scene expression may be further augmented with a global reference term such as $g_{\text{global}}$, yielding the final compositional formula used to instantiate a benchmark prompt.

\textbf{Scope of the formula.}
The proposed formula is not intended to represent the full natural-language prompt. Instead, it only describes the \emph{combination structure of reference content}. We assume that the text-to-image backbone can already handle ordinary textual instructions reasonably well, and therefore focus only on the part of the prompt that involves reference-conditioned content. As a result, pure text-only modifications are not explicitly included in the formula unless they participate in a reference-dependent operation through $g$. For example, ordinary editing instructions such as changing a color are treated as textual modifications rather than part of the formula structure.

\textbf{Structural constraint.}
In practice, the formula structure is deliberately kept simple. The scene representation uses at most two nested levels of the composition operator $C$. The inner level is mainly used to express reference-grounded relational composition, such as cases involving $g_{\mathrm{rel}}$, while the outer level is used to form the complete scene expression. Text-specified relations are omitted from the formula unless they are necessary for disambiguation. As illustrated by the formula below, this assumption is sufficient for the vast majority of benchmark cases while keeping the template space interpretable and manageable.

\begin{equation}
\label{eq:nested_formula}
F =
\underbrace{
  C\Bigl(
    \underbrace{C(f_1,\, T_e \oplus g_1) \oplus g_{\mathrm{rel}}}_{\text{inner composition}},
    \, f_2 \oplus g_2 \oplus g_3
  \Bigr)
}_{\text{outer composition}}
\oplus g_{\mathrm{global}}.
\end{equation}

\textbf{Template construction and sampling.}
Based on the above representation, we sample valid templates across all slot levels under controlled distributions. We explicitly control the proportions of templates containing different numbers of $g_{\mathrm{rel}}$ and $g_{\mathrm{global}}$ terms, preventing the sampled cases from collapsing to a few repeated flat structures while maintaining diverse relational and global-reference patterns.

\textbf{Design goal.}
This design serves two purposes. First, it increases reference-conditioned structural complexity with slot number in a controlled and interpretable way. Second, it prevents the benchmark from being dominated by a narrow set of repeated formula patterns, especially at larger slots. As a result, the final benchmark includes both simple reference transfer cases and more structured multi-reference compositions involving relational constraints and global reference attributes.

\textbf{Representative templates.}
To give a concrete picture of the resulting distribution, Table~\ref{tab:example_formula_by_slot} lists some example formula templates for slots 1--8 after template construction and sampling. As the slot number increases, the dominant patterns gradually shift from simple anchored entities or single-attribute transfer to more complex compositions of the same small set of atomic operators.

\subsection{Prompt Realization Format}
\label{sec:prompt_realization}

\begin{figure}[t]
    \centering
    \includegraphics[width=0.9\linewidth]{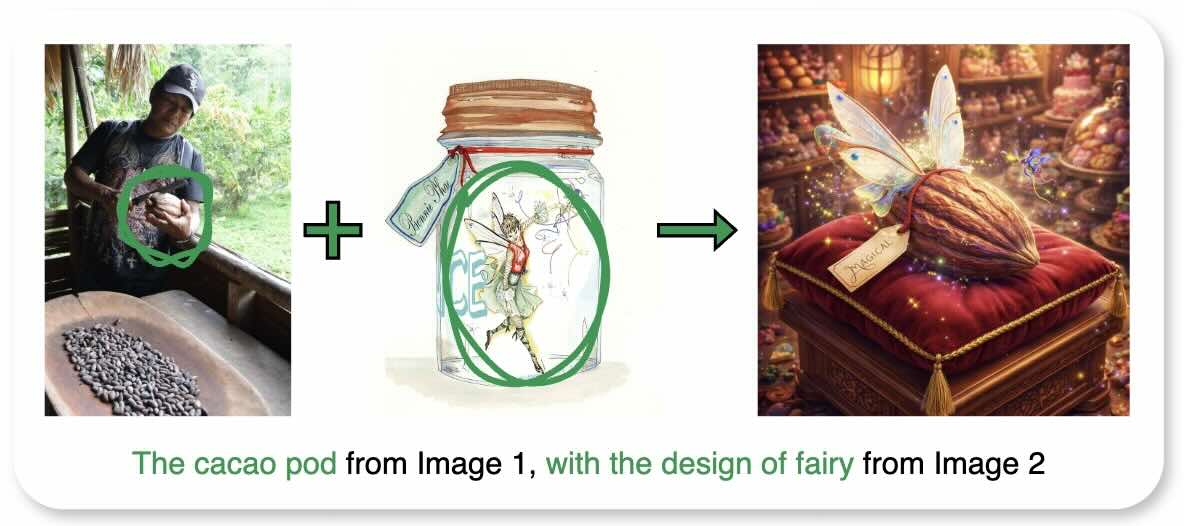}
    \caption{A Cacao-pod Case}
    \label{fig:case_3}
\end{figure}

Given a compositional formula and its associated reference images, we further convert them into a natural-language reference prompt for image generation. This step is particularly important in our setting because many images in the reference pool are visually complex, and the source objects for \textbf{Anchor} or \textbf{Disentangle} often require relatively detailed grounding descriptions. At the same time, we avoid using overly artificial placeholders or highly specialized prompt markup, since such forms may be unnatural for image generation models. We therefore adopt a natural-language prompt realization format that keeps the prompt fluent while preserving explicit source-target grounding.

Our realization format contains two parts. The first paragraph describes the target scene in natural language, where each reference-conditioned target object is introduced as a readable referring expression such as (\texttt{'man\_A'}). The second paragraph explicitly specifies the source-target assignments for all reference operations, including the reference source image, the source content to be extracted, and the target object in the realized prompt. In this way, the main prompt remains natural, while the reference mapping remains explicit. A real example is shown below (corresponding to Fig.~\ref{fig:case_3}).

\begin{lstlisting}[style=promptstyle, caption={Example of natural-language prompt realization.}, label={lst:prompt_realization_example}]
Generate a new scene. In a whimsical fantasy bakery, a special enchanted confection is displayed on a velvet cushion. This item is a unique ('cacao_pod_A'), which looks like a real cacao pod but has been magically altered.

('cacao_pod_A') references [the cacao pod being cut in the man's hands in Image 1].
('cacao_pod_A') references [the fairy in the jar in Image 2].
\end{lstlisting}

\subsection{Checklist Construction from Structured Prompts}
\label{sec:checklist_construction}

Our evaluation checklist is generated from the structured information preserved during prompt realization. When converting a symbolic case into a natural-language reference prompt, we retain the corresponding source--target mappings and operator-level structure. This intermediate representation is then passed to an LLM, which produces checklist questions aligned with the evaluation target of each operator.

The main design principle is that each operator should be evaluated through multiple binary questions rather than a single scalar or holistic judgment. This is necessary because operator-level success is often not atomic. For example, an output may contain the correct target object but fail to match the referenced source faithfully, or an attribute may be transferred but bound to the wrong carrier. A single binary judgment would be too coarse to distinguish such cases, while fully open-ended or non-binary judgments may introduce additional bias and reduce consistency. We therefore decompose the evaluation of each operator into several binary questions, each focusing on one concrete aspect of correctness.

This principle is applied consistently to all four operators, namely $f$, $g$, $\oplus$, and $C$. For an anchored entity $f$, the checklist may separately verify whether the target exists and whether it matches the referenced source entity. For a disentangled attribute $g$, the checklist may separately ask whether the intended attribute appears on the target and whether it is faithful to the reference image. For $\oplus$, multiple questions are used to check whether the transferred attribute is present, whether it is correctly bound to the intended carrier, whether the carrier itself remains coherent, and whether the transferred content resembles the referenced source. For $C$, the checklist likewise separates coexistence, relation satisfaction, and structural coherence into different binary checks. In this way, operator-level failures can be localized more precisely instead of being collapsed into a single judgment.

Each generated question also records the corresponding target part in the structured representation. This design makes the checklist easy to trace back to the original formula and reference prompt, and also supports later grouping and aggregation by operator type, target entity, or failure mode. As a result, the checklist is both fine-grained enough to capture diverse operator-level errors and structured enough to support systematic analysis.

For completeness, we show the full checklist for the cacao-pod case (Fig.~\ref{fig:case_3}) below. Here, the \textit{Target} field records the corresponding operator target or target part in the formula structure, which makes the generated questions easier to trace back to the original prompt representation.

\begin{table*}[!t]
\centering
\footnotesize
\setlength{\tabcolsep}{4pt}
\renewcommand{\arraystretch}{1.08}
\begin{tabularx}{\textwidth}{>{\centering\arraybackslash}p{0.06\textwidth} >{\raggedright\arraybackslash}p{0.22\textwidth} X}
\toprule
\textbf{Op.} & \textbf{Target} & \textbf{Binary question} \\
\midrule
\multicolumn{3}{l}{\textbf{Case formula:} $C(f_1 \oplus g_1)$} \\
\multicolumn{3}{l}{\textbf{Target object:} \texttt{('cacao\_pod\_A')}} \\
\midrule

$f$ & $f_1$ & Does a food item that is clearly a cacao pod, retaining its rugby-ball shape and vertically grooved surface, exist in the generated image? \\
$f$ & $f_1$ & Does \texttt{('cacao\_pod\_A')} match the referenced cacao pod in terms of its core identity as a cacao pod? \\

$g$ & $g_1$, IP on \texttt{cacao\_pod\_A} & Does \texttt{('cacao\_pod\_A')} incorporate the overall fairy-like design, including wings, pose, and magical effects? \\
$g$ & $g_1$, IP on \texttt{cacao\_pod\_A} & Are recognizable fairy-design cues present in the overall design of \texttt{('cacao\_pod\_A')}? \\
$g$ & $g_1$, IP on \texttt{cacao\_pod\_A} & Is the color scheme and motif of the referenced fairy transferred to the overall design of \texttt{('cacao\_pod\_A')}? \\
$g$ & $g_1$, IP on \texttt{cacao\_pod\_A} & Is the overall fairy design integrated with \texttt{('cacao\_pod\_A')} while preserving its identity as a cacao pod? \\

$\oplus$ & $\oplus,\ f_1 \oplus g_1$, carrier & After the fusion, is \texttt{('cacao\_pod\_A')} still clearly recognizable as itself and structurally intact? \\
$\oplus$ & $\oplus,\ f_1 \oplus g_1$, fit & Is the transferred IP clearly visible on \texttt{('cacao\_pod\_A')} and naturally integrated? \\
$\oplus$ & $\oplus,\ f_1 \oplus g_1$, fit & Does the fusion remain visually coherent and physically plausible on \texttt{('cacao\_pod\_A')}? \\
$\oplus$ & $\oplus,\ f_1 \oplus g_1$, exclusivity & Is the transferred IP confined to the intended scope (only on $f_1$), without leaking to other entities or the background? \\

$C$ & $C$, spatial & Is the scene composition spatially coherent and physically plausible? \\
$C$ & $C$, relation & Does the generated image satisfy the intended scene relation? \\
$C$ & $C$, duplication & Does any prompted entity appear more times than intended in the generated image? \\
$C$ & $C$, leakage & Does any unintended source content appear in the generated image beyond the referenced cacao-pod content? \\
$C$ & $C$, leakage & Does any unintended source content appear in the generated image beyond the intended fairy-design transfer on \texttt{('cacao\_pod\_A')}? \\

\bottomrule
\end{tabularx}
\caption{Full checklist example for the cacao-pod case. The \textit{Target} field records the corresponding operator target or target part in the formula structure, which supports tracing and later grouping.}
\label{tab:full_checklist_cacao}
\end{table*}

\section{Evaluation and Diagnostic Details}
\subsection{Reliability of the Evaluation Protocol}
\label{sec:eval_reliability}
To assess whether our evaluation depends on the choice of VLM judge, we conduct a human audit involving Gemini-2.5-Pro (G25P), Gemini-3-Pro (G3P), GPT-5.1, and GPT-5.4. We sample 200 benchmark cases and generate each case with both Nano Banana 2 and Emu3.5, yielding 400 outputs. For each output, human annotators answer the same binary checklist questions used by the VLM judges. We aggregate the checklist decisions associated with each operator and normalize the resulting operator-level scores. Pearson and Spearman measure linear and rank correlation, respectively, between VLM and human operator scores; MAE measures their normalized score difference, while agreement is the percentage of individual checklist decisions that match the human annotations.

\begin{table}[t]
\centering
\caption{Alignment between VLM judges and human annotations on outputs from 200 sampled benchmark cases. The ensemble averages the four VLM judges.}
\label{tab:judge_human_alignment}
\small
\setlength{\tabcolsep}{3.5pt}
\begin{tabular}{lcccc}
\toprule
\textbf{Judge} & \textbf{Pear.}$\uparrow$ & \textbf{Spear.}$\uparrow$ & \textbf{MAE}$\downarrow$ & \textbf{Agree.}$\uparrow$ \\
\midrule
G3P     & 0.608 & 0.613 & 0.152 & 86.8\% \\
G25P    & 0.554 & 0.537 & 0.173 & 85.4\% \\
GPT-5.1 & 0.569 & 0.558 & 0.162 & 88.1\% \\
GPT-5.4 & 0.580 & 0.542 & 0.156 & 88.4\% \\
Ensemble & \textbf{0.662} & 0.604 & 0.153 & \textbf{88.6\%} \\
\bottomrule
\end{tabular}
\end{table}

As shown in Table~\ref{tab:judge_human_alignment}, all four VLM judges achieve 85.4--88.4\% checklist-level agreement with human annotations and exhibit broadly comparable operator-level correlations and errors. This indicates that the checklist-based evaluation is not tied to a single judge. We use G25P for full-benchmark evaluation because it provides a practical trade-off between reliability and evaluation cost. The ensemble further improves Pearson correlation and agreement, providing a higher-confidence option when additional evaluation cost is acceptable.

\subsection{Complete Diagnostic Tree Rules}
\label{sec:diagnostic_tree_rules}

The diagnostic tree is used to localize the source of failure in a complex multi-reference case. Starting from the full formula at the root node, we recursively simplify the case into a set of easier sub-cases and compare the model behavior across nodes.

\textbf{General principle.}
Each child node should preserve the same overall scene as the root case, while reducing part of the reference-conditioned complexity. Removed reference content is not simply deleted; when necessary, it is downgraded to an ordinary text-only description so that the scene context remains comparable across nodes.

\textbf{Rule 1: Global-reference stripping.}
If the formula contains one or more \(g_{\mathrm{global}}\) terms, we remove them one by one. This rule is used to diagnose whether global references, such as style or scene-level constraints, interfere with lower-level anchor or attribute fidelity.

\textbf{Rule 2: Composition flattening.}
If the formula contains a composition operator \(C(\cdot)\), we flatten it into simpler branches. One child node keeps one reference-conditioned branch, while the remaining branches are downgraded to text-only scene descriptions. This rule is especially useful for identifying whether a failure only appears under joint composition. In particular, when multiple anchors co-occur in the same scene, composition flattening can isolate one anchor at a time while retaining the rest of the scene as ordinary textual context.

\begin{figure}[t]
    \centering
    \includegraphics[width=0.85\linewidth]{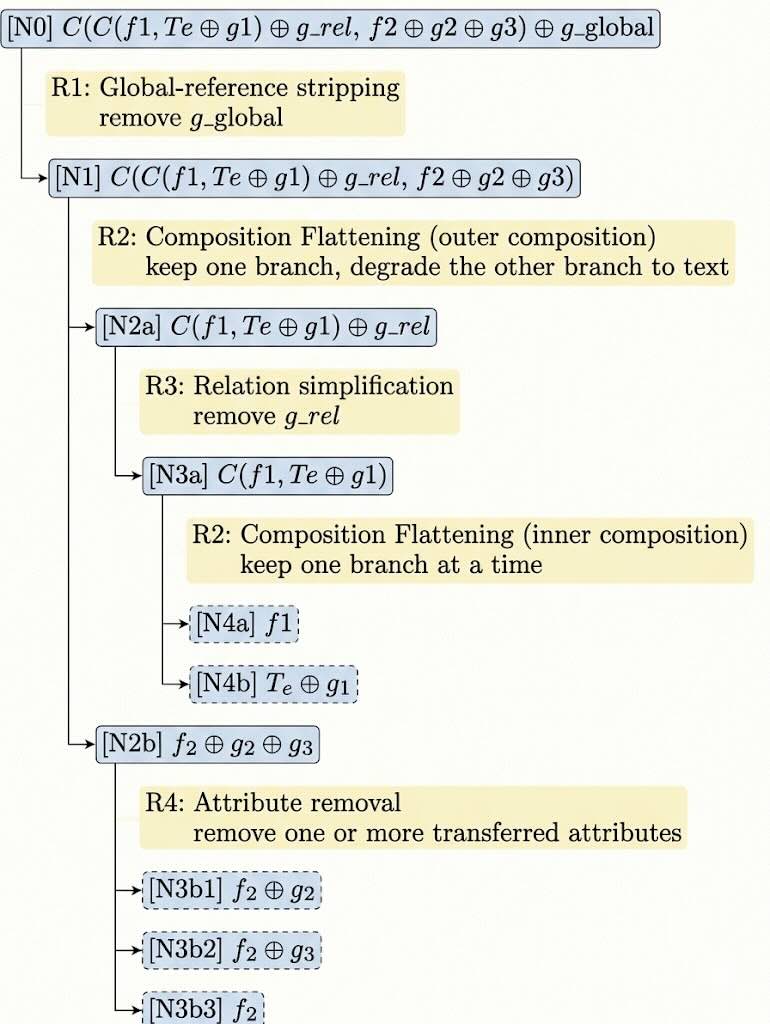}
    \caption{A formula-level example of diagnostic tree decomposition.}
    \label{fig:diagnostic_tree_rules_example}
\end{figure}

\textbf{Rule 3: Relation simplification.}
If a node contains an explicit interaction term, the relation is simplified before removing the participating branches themselves. In the simplified child node, the original interaction is converted into a text-only relation description while the main scene context is preserved. This helps distinguish failures caused by relation grounding from failures caused by entity appearance or attribute transfer.

\textbf{Rule 4: Attribute removal.}
For expressions of the form \(E \oplus g\), \(E \oplus g_{\mathrm{ip}}\), or \(E \oplus g_{\mathrm{attach}}\), we may remove the added attribute while preserving the carrier entity \(E\). This rule is used to determine whether the failure comes from the carrier anchor itself or from the added attribute transfer.

\textbf{Stopping criterion.}
The decomposition stops when the remaining node contains only a single informative reference-conditioned unit, or when further simplification would no longer help isolate a more specific source of failure. In practice, leaf nodes usually correspond to a single anchor, a single attribute transfer, or a minimally composed scene.

\textbf{Node-to-question mapping.}
Each diagnostic node is evaluated only with the checklist items that correspond to the retained operator targets in that node. Therefore, the diagnostic tree is not only a formula decomposition, but also an evaluation decomposition. The root node uses the full checklist of the original case, while each child node uses the subset of questions that remains relevant after simplification.

Figure~\ref{fig:diagnostic_tree_rules_example} shows a formula-level example of diagnostic tree decomposition. Starting from the full case at \(N_0\), we first apply Rule~1 to strip the global reference term \(g_{\mathrm{global}}\), yielding \(N_1\). We then apply Rule~2 to flatten the outer composition, which separates the left relational subscene \(N_{2a}\) from the right attribute-transfer branch \(N_{2b}\). On the left branch, Rule~3 removes the relation term \(g_{\mathrm{rel}}\), and Rule~2 is applied again to flatten the remaining inner composition into two simpler nodes, \(N_{4a}\) and \(N_{4b}\). On the right branch, Rule~4 removes one or more transferred attributes, producing the simplified nodes \(N_{3b1}\), \(N_{3b2}\), and \(N_{3b3}\). This example illustrates how the complete rule set recursively reduces a complex formula into a set of simpler diagnostic branches while preserving the same overall scene context.

\subsection{Quantitative Validation of Diagnostic Trees}
To validate diagnostic reliability, we construct diagnostic trees for 200 Emu3.5 cases and generate an image at every node. A VLM evaluates each node, after which the rules above localize the source of each operator-level failure. Independently, human annotators inspect the same trees and identify the node at which each failure originates. The automatic and human localizations agree in \textbf{82.6\%} of cases.

Table~\ref{tab:diagnostic_sources} in the main paper summarizes the localized failure sources. Here, we additionally examine where requirements become solvable along the simplification process. For each root-to-leaf path, we normalize node position as \emph{relative depth}, with 0 denoting the original complete case and 1 the maximally simplified node. For each operator requirement, we record the earliest depth at which its evaluation changes from failure to success. The cumulative pass rate at a given depth is the proportion of requirements that are already successful at the root or first become successful by that point. Requirements that remain unsuccessful at every node are treated as persistent failures and therefore do not enter the cumulative count.

\begin{figure}[t]
\centering
\includegraphics[width=0.9\linewidth]{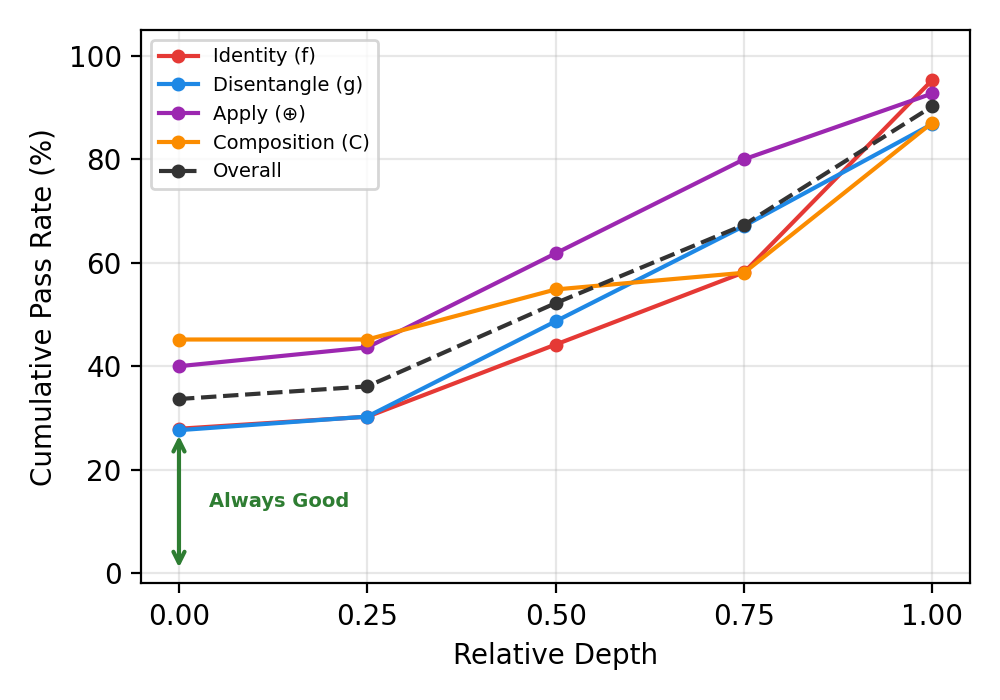}
\caption{Cumulative pass rate over relative diagnostic-tree depth on 200 Emu3.5 cases.}
\label{fig:diagnostic_quantitative}
\end{figure}

As shown in Fig.~\ref{fig:diagnostic_quantitative}, the cumulative pass rate rises consistently with relative depth for all four operators. Thus, many requirements that fail in the complete case become solvable only after interfering reference-conditioned components are removed. Conversely, the endpoints remain below 100\% because some requirements persistently fail even in the simplest diagnostic nodes. Together with the source distribution in Table~\ref{tab:diagnostic_sources}, this result shows that many observed failures arise from interactions introduced by more complex formula structure, while a smaller subset reflects difficulty intrinsic to the isolated reference unit.

\begin{figure}[!h]
    \centering
    \includegraphics[width=0.999\linewidth]{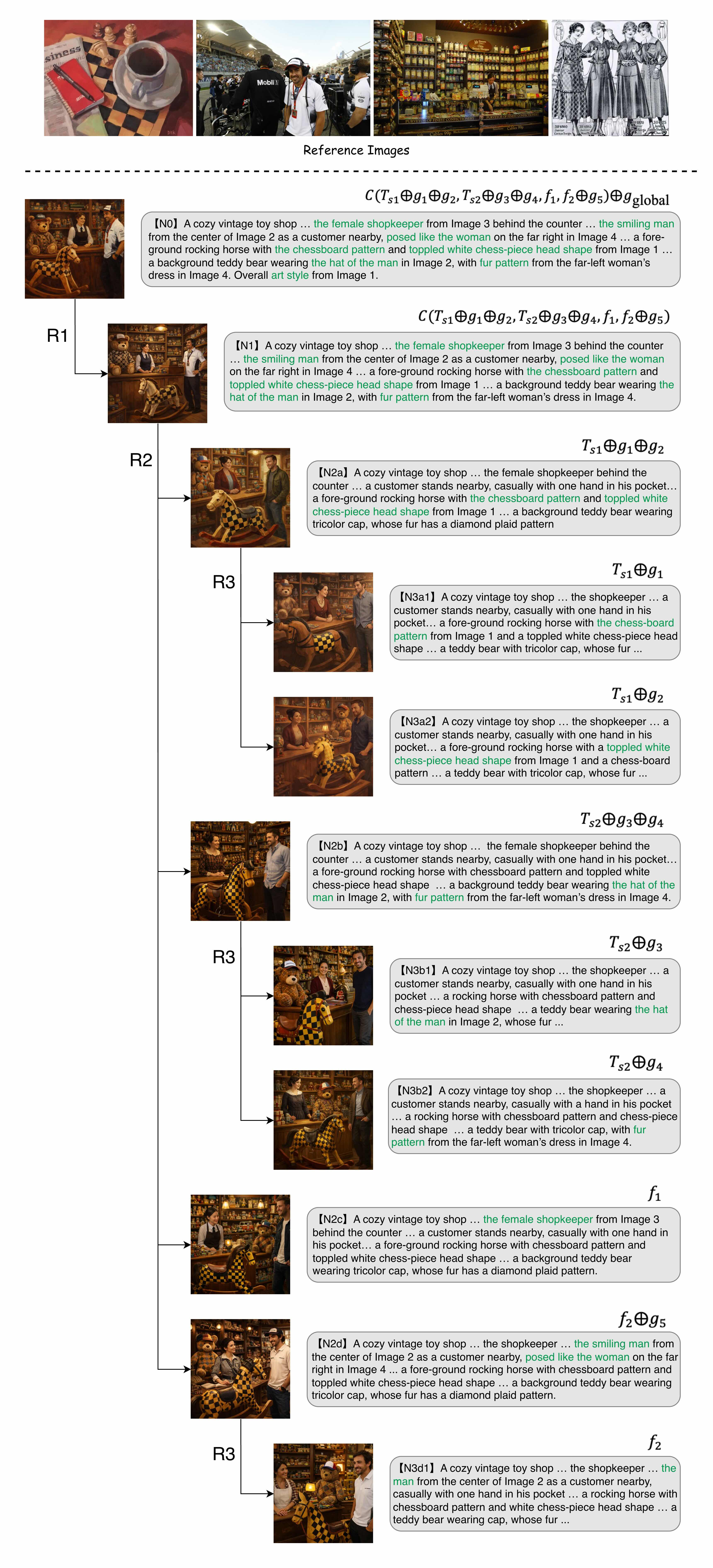}
    \caption{A diagnostic tree example for a complex multi-reference case. Starting from the full case at \(N_0\), the tree is expanded by sequentially applying global-reference stripping, composition flattening, relation simplification, and attribute removal. The resulting branches isolate different potential sources of failure in a structured way.}
    \label{fig:diagnostic_tree_example}
\end{figure}

\subsection{Additional Diagnostic Tree Examples}

Figure~\ref{fig:diagnostic_tree_example} shows an additional diagnostic tree example, corresponding to the third-from-last row in Fig.~\ref{fig:more_case}. In the root node \(N_0\), GPT-Image-1.5 shows two visible problems. First, the chess-piece-inspired head shape of \((\texttt{horse\_A})\) does not appear at all. Second, the hat on \((\texttt{bear\_A})\) is present, but its color does not faithfully match the referenced hat. The diagnostic tree helps determine whether these errors are caused by the same underlying source.

For the horse branch, the chess-piece-inspired head shape never appears from the root node down to \(N_{3a2}\). This indicates that GPT-Image-1.5 does not reliably realize this shape-related reference requirement itself. In other words, the disentangle-and-apply process for this structural attribute is already failing even after the case is simplified, rather than the error being introduced only by additional scene complexity.

The bear branch exhibits a different pattern. In \(N_0\), the bear is generated, but the transferred hat attribute is not faithful to the reference, since its color is mismatched. In \(N_{2b}\), however, the failure changes form: the bear itself is not generated, so the error is no longer only about hat fidelity, but about the carrier entity collapsing altogether. By contrast, the hat becomes much more accurate in \(N_1\) and \(N_{3b1}\). This suggests that the model is not uniformly incapable of realizing the hat transfer; instead, the corresponding content is preserved unstably, and the failure mode changes across compositional contexts.

Overall, this example shows that the diagnostic tree can distinguish between two qualitatively different situations: a reference requirement that is consistently not realized at all, and a reference requirement whose behavior is unstable, with the observed error shifting between attribute mismatch and carrier-level failure.

\begin{figure}[!h]
    \centering
    \includegraphics[width=0.75\linewidth]{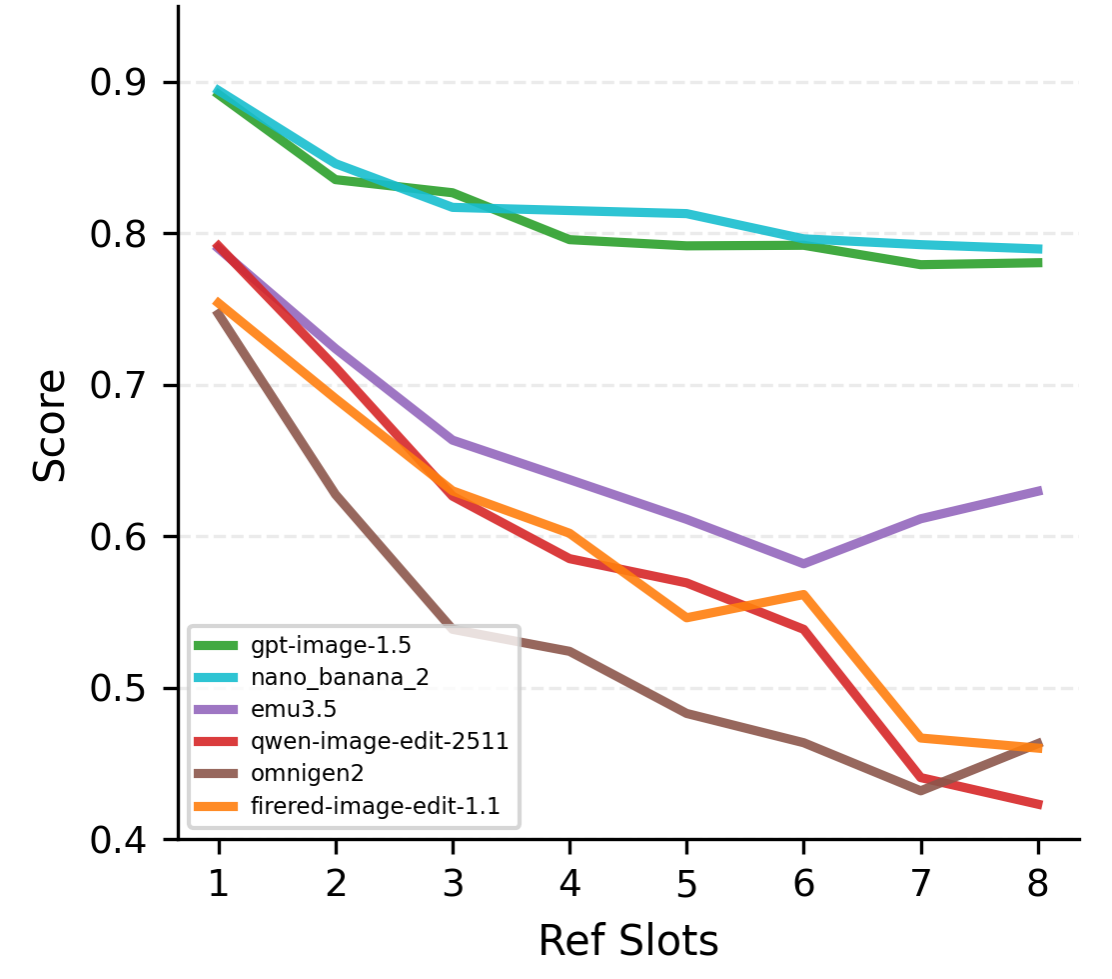}
    \caption{Overall operator-aligned performance across slot levels.}
    \label{fig:slot_performance}
\end{figure}

\section{Additional Experimental Results}
\subsection{Performance across Slot Levels}

Figure~\ref{fig:slot_performance} reports average operator-aligned performance from slot 1 to slot 8. Performance generally declines as slot count increases, supporting its use as a controllable measure of formula structure. The decline is substantially sharper for open-source models, whereas the leading proprietary models remain comparatively stable as more reference-conditioned elements are introduced. The trend is not strictly monotonic for every model because slot count does not fully determine empirical difficulty, which also depends on content-level factors such as reference clutter, entity composition, and attribute granularity.

\begin{table*}[!t]
\centering
\small
\setlength{\tabcolsep}{6pt}
\renewcommand{\arraystretch}{1.12}
\begin{tabular}{lll}
\toprule
\textbf{Benchmark / application task} & \textbf{Typical requirement} & \textbf{TRACE-Bench abstraction} \\
\midrule
MultiBanana: X Objects & multiple referenced objects & $C(f_1, f_2, \ldots, f_X)$ \\
MultiBanana: X--1 Objects + Local & objects + one local attribute reference & $C(f_1, \ldots, f_{X-1} \oplus g_{\text{local}})$ \\
MultiBanana: X--1 Objects + Global & objects + one global reference & $C(f_1, \ldots, f_{X-1}) \oplus g_{\text{global}}$ \\
MultiBanana: X--1 Objects + Background & objects + one background reference & $C(f_1, \ldots, f_{X-1}) \oplus g_{\text{background}}$ \\
\midrule
MICON: Object Composition & combine multiple referenced instances & $C(f_1, f_2, \ldots, f_n)$ \\
MICON: Spatial Composition & compose multiple instances with relation & $C(f_1, f_2, \ldots, f_n) \oplus T_{\text{rel}}$ \\
MICON: Attribute Disentanglement & transfer one disentangled attribute & $T_e \oplus g$ \quad or \quad $f \oplus g$ \\
MICON: Component Transfer & transfer an attachable component & $T_e \oplus g_{\text{attach}}$ \quad or \quad $f \oplus g_{\text{attach}}$ \\
MICON: FG/BG Composition & combine foreground and background references & $C(f_1, \ldots, f_n) \oplus g_{\text{background}}$ \\
MICON: Story Generation & multi-entity narrative scene generation & $C(f_1, \ldots, f_n) \oplus T_{\mathrm{rel}}$ \\
\midrule
Application: Virtual Try-On & transfer garments or accessories to a target person & $f(\texttt{person}) \oplus g_{\text{attach},1} \oplus g_{\text{attach},2} \oplus \cdots$ \\
Application: Group Photo Layout & compose multiple subjects under a layout constraint & $C(f_1, f_2, \ldots, f_n) \oplus g_{\text{layout}}$ \\
Application: Novel View Synthesis & preserve the same subject under a changed viewpoint & $f \oplus g_{\text{view}}$ \\
Application: IP-style Reference & transfer overall design language to a target carrier & $T_e \oplus g_{\text{ip}}$ \quad or \quad $f \oplus g_{\text{ip}}$ \\
Application: Stylization & preserve scene content under a global style transfer & $C(f_1, \ldots, f_n) \oplus g_{\text{style}}$ \\
\bottomrule
\end{tabular}
\caption{Examples of mapping benchmark-defined and application-oriented tasks into the TRACE-Bench formula space.}
\label{tab:application_formula_mapping}
\end{table*}

\subsection{Cross-Model Qualitative Comparisons}

\begin{figure*}[h]
    \centering
    \includegraphics[width=0.96\linewidth]{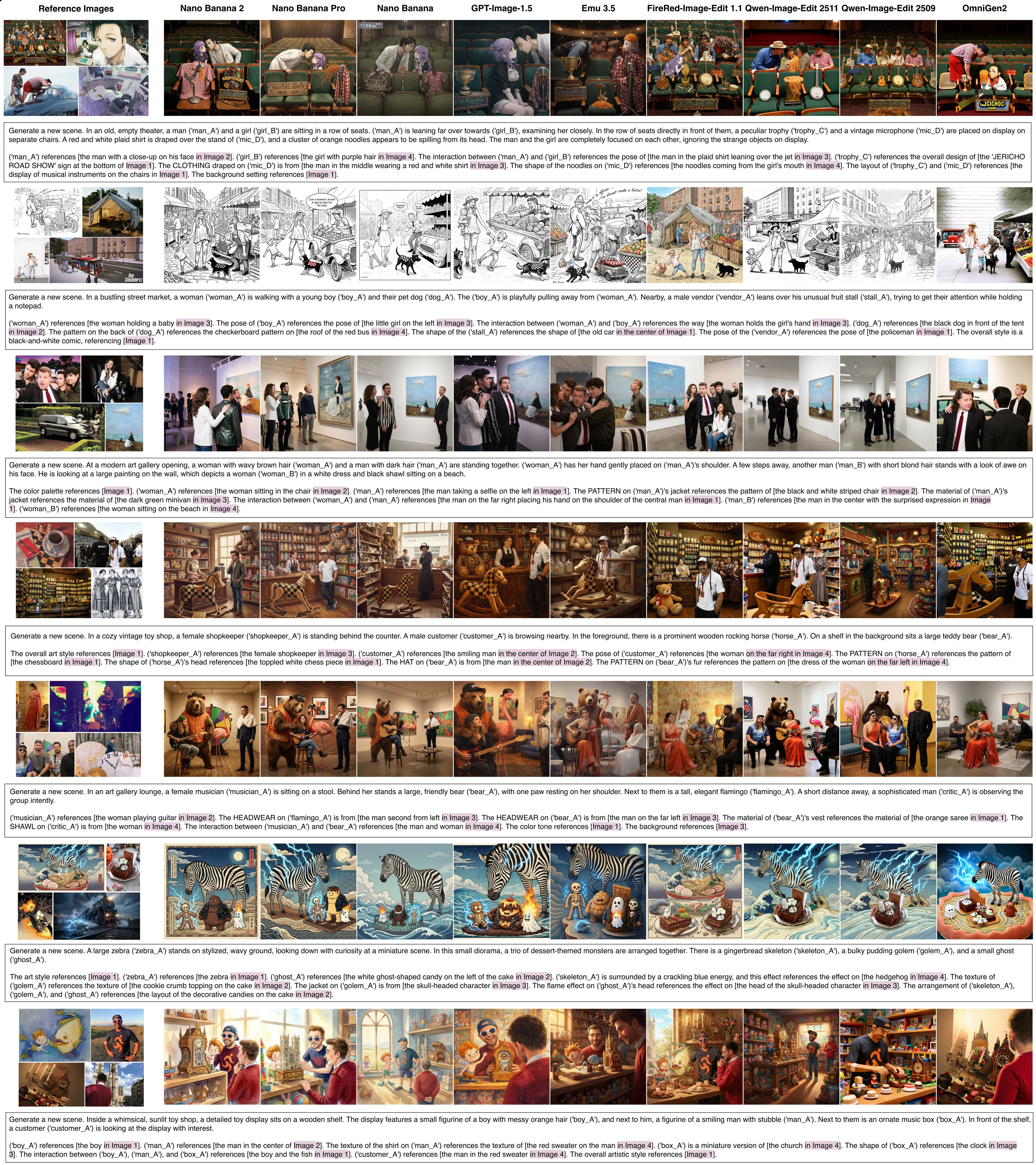}
    \caption{Additional cross-model qualitative examples from TRACE-Bench. Each row presents one benchmark case, including the reference images, the realized natural-language reference prompt, and outputs from multiple representative models. These examples highlight recurring differences in reference retention, compositional fidelity, and stylistic tendencies across models, especially under complex multi-reference settings.}
    \label{fig:more_case}
\end{figure*}

Figure~\ref{fig:more_case} provides additional cross-model qualitative comparisons on several representative cases from TRACE-Bench. Several consistent patterns can be observed. First, under high-complexity settings such as slot-8 cases, open-source models are often able to retain multiple reference-conditioned contents simultaneously, whereas closed-source models more often drop part of the reference information and instead fall back to generic text-to-image generation. Second, among the open-source models, Emu3.5 tends to preserve more reference content overall, while FireRed occasionally produces unusual blurring artifacts in slot-8 cases. We also observe that Qwen-Image-Edit-2509 and Qwen-Image-Edit-2511 generate highly similar outputs in certain cases, suggesting closely related generation behavior. Finally, across closed-source models, the GPT and Gemini families exhibit noticeably different image-generation tendencies and stylistic preferences, even when given the same reference prompt.

\subsection{Performance across Attribute Subtypes}

\begin{figure}[!h]
    \centering
    \includegraphics[width=0.9\linewidth]{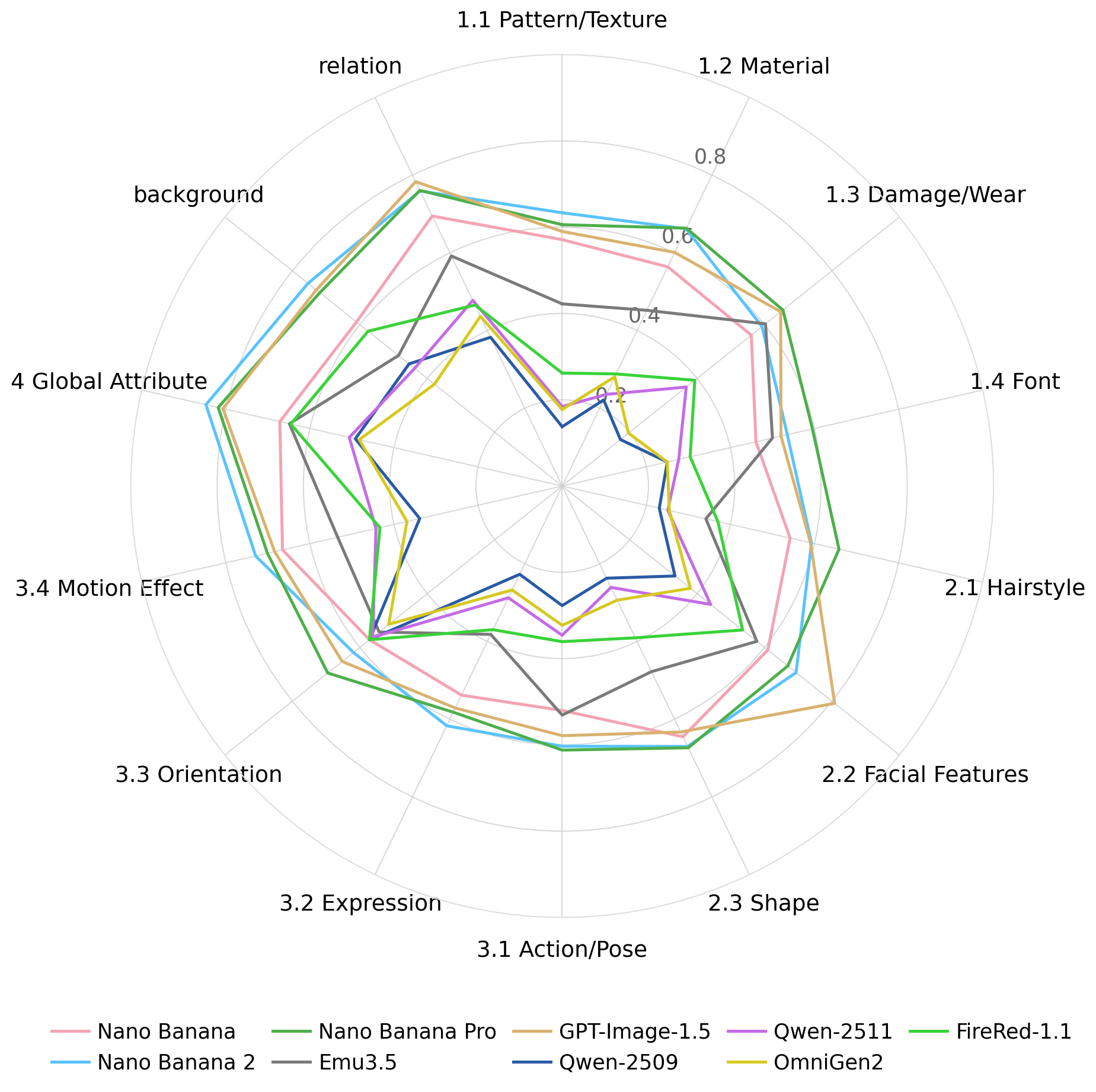}
    \caption{Breakdown of disentangle performance across fine-grained attribute subtypes. Each axis corresponds to one subtype in the tagging taxonomy.}
    \label{fig:g_subtype_radar}
\end{figure}

Since the disentangle operator covers a diverse set of attribute types, an overall $g$ score may hide important differences across sub-capabilities. We therefore further analyze model performance by attribute subtype, as shown in Fig.~\ref{fig:g_subtype_radar}.

\subsection{More Application-Oriented Formula Abstractions}
\label{sec:application_oriented_formulas}

To further illustrate the practical coverage of our formulation, Table~\ref{tab:application_formula_mapping} maps representative task categories from existing benchmarks, together with several common application-oriented settings, into the TRACE-Bench formula space. 

The key point is that diverse and realistic task categories can be expressed within the same formula space by assigning different types of reference-conditioned content to the same compositional structure. In this way, our abstraction is not limited to benchmark-specific categories, but can also cover a broad range of practical generation settings within an operator-aligned framework. Moreover, by combining formula-level sampling with instance- and attribute-level sampling, our framework also has the potential to support large-scale generation of diverse multi-reference tasks in a systematic way.

\section{Potential Extensions}
\label{sec:potential_extensions}

TRACE-Bench currently focuses on diagnosing multi-reference image generation, while its operator formulation also suggests several concrete extensions that preserve the alignment between case construction, operator targets, and diagnostic questions. At the model level, Anchor could combine decoupled image-prompt attention with query-conditioned localization to retain reference-specific evidence while suppressing salient distractors~\cite{ye2023ipadapter,liu2023autr}. Disentangle and Apply could combine feature-level comparison and prototype memories with localized attention that reduces identity mixing when several references must be bound to distinct targets~\cite{liu2022tis,ye2021uda,xiao2023fastcomposer}. Future benchmark annotations could further include promptable region masks for individual operator instances, optionally bootstrapped with diffusion-derived pseudo-masks or direct mask generation~\cite{kirillov2023segment,ma2023diffusionseg,ma2025freesegdiff,yang2026genmask}; these localized targets would allow diagnostic child cases to provide structured feedback for reward-guided model updates, iterative refinement, or evidence-based revisiting of an initial diagnosis~\cite{xu2023imagereward,ma2021bsiris,ma2026rere}. For benchmark construction, detector-verifiable properties such as object co-occurrence, position, count, and color~\cite{ghosh2023geneval} could provide scalable checks for newly sampled formula templates. Future releases could treat rare concepts and underrepresented combinations of operators and attributes as explicit difficulty axes and expand them through generation with quality filtering~\cite{oshima2025multibanana,zhang2024rareness}; they could also reduce annotation costs through scalable multimodal data curation and composition of compatible labeled resources~\cite{gadre2023datacomp,liu2023annotationfree}. Beyond the current image setting, a video extension could ground reference evidence to query-conditioned temporal moments and derive operator-level supervision from partial temporal annotations and progressive pseudo-label refinement~\cite{lei2021qvhighlights,ju2023constraintunion,wang2025contrastunity}. Promptable masks could provide localized reference tracks across frames, while efficient spatiotemporal adaptation could support longer reference sequences~\cite{ravi2025sam2,yang2025moma}. Its evaluation could combine intermediate state, motion, contact, and temporal-order checks with video-specific dimensions such as subject consistency, motion smoothness, and temporal flickering, rather than relying on final-frame quality alone~\cite{xia2026roboprocessbench,huang2024vbench}. A separate cross-view variant could instantiate $g_{\text{view}}$ through radiance-field view synthesis and depth-aware generalizable rendering, testing whether reference identity and geometry remain consistent across viewpoint changes~\cite{mildenhall2020nerf,shi2022darf}. Together, these directions would preserve the central principle of TRACE-Bench by ensuring that each added capability remains explicit in case construction and independently diagnosable during evaluation.

\clearpage

\end{document}